\documentclass{article}
\usepackage[preprint]{neurips2025}

\usepackage{graphicx}
\usepackage{subcaption}
\usepackage{algorithm} 
\usepackage{algpseudocode} 
\usepackage{import}
\usepackage{amsmath}
\usepackage{amsthm}
\usepackage{xcolor}
\usepackage{hyperref} 
\usepackage{lipsum}
\usepackage{enumitem,kantlipsum}
\usepackage{cases}
\setlist{topsep=0pt, leftmargin=!}

\usepackage{caption} 
\usepackage{pifont}
\usepackage{amsmath,amsfonts,bm}

\def\eqref#1{equation~\ref{#1}}

\def\1{\bm{1}}

\def\eps{{\epsilon}}
\def\veps{{\bm{\epsilon}}}

\def\vmu{{\bm{\mu}}}
\def\vtheta{{\bm{\theta}}}
\def\va{{\bm{a}}}
\def\vb{{\bm{b}}}

\def\ve{{\bm{e}}}
\def\vf{{\bm{f}}}

\def\vu{{\bm{u}}}
\def\vv{{\bm{v}}}
\def\vw{{\bm{w}}}
\def\vx{{\bm{x}}}
\def\vy{{\bm{y}}}
\def\vz{{\bm{z}}}

\def\mA{{\bm{A}}}

\def\mC{{\bm{C}}}
\def\mD{{\bm{D}}}

\def\mF{{\bm{F}}}

\def\mI{{\bm{I}}}

\def\mM{{\bm{M}}}

\def\mP{{\bm{P}}}
\def\mQ{{\bm{Q}}}
\def\mR{{\bm{R}}}

\def\mT{{\bm{T}}}

\def\mPhi{{\bm{\Phi}}}

\def\vphi{{\bm{\phi}}}
\def\mPi{{\bm{\Pi}}}

\def\vpi{{\bm{\pi}}}
\def\vmu{{\bm{\mu}}}

\DeclareMathAlphabet{\mathsfit}{\encodingdefault}{\sfdefault}{m}{sl}
\SetMathAlphabet{\mathsfit}{bold}{\encodingdefault}{\sfdefault}{bx}{n}

\def\gA{{\mathcal{A}}}
\def\gB{{\mathcal{B}}}
\def\gC{{\mathcal{C}}}
\def\gD{{\mathcal{D}}}

\def\gF{{\mathcal{F}}}
\def\gG{{\mathcal{G}}}

\def\gI{{\mathcal{I}}}

\def\gM{{\mathcal{M}}}

\def\gP{{\mathcal{P}}}

\def\gS{{\mathcal{S}}}

\def\sC{{\mathbb{C}}}

\def\sN{{\mathbb{N}}}

\newcommand{\E}{\mathbb{E}}

\newcommand{\R}{\mathbb{R}}

\DeclareMathOperator*{\argmax}{arg\,max}

\newtheorem{theorem}{Theorem}[section]
\newtheorem{assumption}[theorem]{Assumption}
\newtheorem{lemma}[theorem]{Lemma}
\newtheorem{definition}[theorem]{Definition}
\newtheorem{remark}[theorem]{Remark}

\newtheorem{proposition}[theorem]{Proposition}
\newtheorem{example}[theorem]{Example}

\title{Understanding the theoretical properties of projected Bellman equation, linear Q-learning, and approximate value iteration}
\author{Han-Dong Lim \\ limaries30@kaist.ac.kr\And Donghwan Lee\\ donghwan@kaist.ac.kr}

\begin{document}

\maketitle
\begin{abstract}
In this paper, we study the theoretical properties of the projected Bellman equation (PBE) and two algorithms to solve this equation: linear Q-learning and approximate value iteration (AVI). We consider two sufficient conditions for the existence of a solution to PBE : strictly negatively row dominating diagonal (SNRDD) assumption and a condition motivated by the convergence of AVI. The SNRDD assumption also ensures the convergence of linear Q-learning, and its relationship with the convergence of AVI is examined. Lastly, several interesting observations on the solution of PBE are provided when using $\eps$-greedy policy.
\end{abstract}
\section{Introduction}

Reinforcement learning (RL) has achieved significant success, exemplified by the deep Q-network (DQN)~\citep{mnih2015human}. This success can be largely attributed to two algorithms: Q-learning~\citep{watkins1992q} and the approximate value iteration (AVI)~\citep{bertsekas2011approximate}. Understanding the behavior of these algorithms has been a central focus of extensive research.

Q-learning, initially developed by~\cite{watkins1992q} in a tabular setup where $Q$-values are stored for every state-action pair, has since been the subject of considerable investigation. Both asymptotic and non-asymptotic analysis of the algorithm have been thoroughly explored in works such as~\citep{borkar2000ode,even2003learning,lee2019unified,chen2022finite,lee2024final}, to list a few.

Moving beyond the tabular setup, function approximation is commonly used to address the problem of large state-action spaces in practical scenarios. Specifically, we focus on the simplest form of approximation: the linear function approximation scheme. However, introducing function approximation brings several challenges. In the case of Q-learning with linear function approximation—referred to as linear Q-learning—two major issues arise: 1) the existence of a solution to the projected Bellman equation (PBE) that the algorithm aims to solve, and 2) the stability of the algorithm. While recent works have explored these challenges~\citep{melo2007convergence,meyn2024projected}, there remains significant opportunity for further advancing our understanding in this area.

Meanwhile, value iteration is one of the simplest algorithms in RL when the model is known. By incorporating linear function approximation into the value iteration framework, the approximate value iteration (AVI) scheme has been widely used~\cite{munos2007performance,mann2014scaling}. AVI also seeks to solve the PBE as linear Q-learning does. Nonetheless, it is not well understood, when the AVI algorithm converges while linear Q-learning does not, and vice versa. 

Overall, the theoretical understanding of PBE and its related algorithms, specifically Q-learning and AVI, are not well-understood. This motivates our study, and the purpose of this paper is to extend our knowledge on these subjects. The main contributions of this study are outlined in the following: 
\begin{enumerate}[wide, labelwidth=!, labelindent=0pt]
  \item A sufficient condition for existence and uniqueness of a solution to PBE:
        \begin{itemize}
            \item We provide a thorough investigation of the existence and uniqueness of a solution to PBE under the assumption of a matrix having strictly negatively row dominating diagonals (SNRDD assumption), which is new in the literature. This assumption includes a wide class of settings: tabular and linear function approximation (with regularization). Moreover, our analysis considers various behavior and target policy scenarios including continuous or lispchitz policies.
            \item A sufficient condition, derived from the AVI framework, is provided. We then explore its relationship to the SNRDD assumption, demonstrating that while the two are generally different, they can coincide under specific additional conditions.
        \end{itemize}
    \item We provide a new convergence proofs for a family of Q-learning algorithms and AVI algorithm, respectively. Furthermore, we provide examples where AVI converges while linear Q-learning does not, and vice-versa.
    \begin{itemize}
        \item   The proof of Q-learning relies on ODE arguments based on contraction theory~\citep{lohmiller1998contraction} and the SNRDD assumption. This covers asynchronous tabular Q-learning, linear Q-learning with SNRDD assumption, and regularized Q-learning~\citep{lim2024regularized}. To author's best knowledge, SNRDD is the first assumption that can be applied to prove convergence of both linear and tabular Q-learning in a unified way using a fixed behavior policy. Regarding regularized Q-learning, the existing assumptions on positiveness and orthogonality of feature matrix are relaxed. 
    
    \item We provide an example showing that, even though the SNRDD assumption ensures the convergence of Q-learning and the existence of a solution to the PBE, the resulting solution may still lead to a sub-optimal policy.

    \item The convergence of AVI follows from the condition that guarantees existence and uniqueness of a solution to PBE. 
    \end{itemize}

        \item  Lastly, we provide two examples explaining the theoretical properties of solutions to PBE when $\eps$-greedy policy is used, which is not covered by previous analysis due to its discontinuity. The first example shows depending on the value of $\eps$, there is a chance of non-existence or multiplicity of the solution even though SNRDD condition is met. The second example shows a pathological phenomenon using $\eps$-greedy policy that increasing the parameter $\eps$ can ensure a solution to PBE that yields an optimal policy, to which Q-learning cannot converge.
\end{enumerate}

\textbf{Related Works:} ~\cite{melo2007convergence,chen2022finite} studied the convergence of linear Q-learning with additional assumptions that might not be satisfied in the tabular setting.~\cite{meyn2024projected,liu2025linear} considered using a version of $\eps$-softmax behavior policy, the so-called tamed-Gibbs policy, and established results that there exists a solution of PBE, and the learning parameters of Q-learning remain bounded. Nonetheless, the tamed-Gibbs policy requires several specific design choices. In contrast, we consider a different scenario and proof approach: existence of the solution is explored for continuous or lipschitz policy under the assumption of SNRDD. In proving the convergence of Q-learning, we consider an arbitrary fixed behavior policy, which is idealistic but different scenario, and this naturally extends the proof idea of Q-learning in the tabular setup. The proof relies on contraction theory~\citep{lohmiller1998contraction}, and its connection offers new insights.

~\cite{lim2024regularized} studied Q-learning with an additional term that serves a similar role to \( l_2 \)-regularization, referred to as regularized Q-learning. Under additional assumptions on the feature matrix, this ensures convergence to a unique point.~\cite{zhang2021breaking} studied Q-learning using target network, projection and regularization. We show that target network, projection or any additional assumptions on feature matrix are not required to prove the convergence of regularized Q-learning.

Several studies have proposed variations of linear Q-learning~\citep{chen2023target,maei2010toward,devraj2017zap,carvalho2020new} which are summarized in the Appendix~\ref{app:sec:related-works}. Although these methods ensure boundedness or convergence, the exact points to which the algorithm converges remain not well understood.

The AVI scheme has been widely studied to tackle the challenges posed by large state-action spaces~\citep{bertsekas1997nonlinear}. Recent research has provided insights into the convergence properties of AVI, highlighting its close connection to algorithms that employ target network updates—a methodology inspired by the success of DQN. ~\cite{lee2020periodic} explored Q-learning in a tabular setting, while ~\cite{asadi2023td,fellows2023target,chetarget2024} investigated temporal difference (TD) learning with target network updates, demonstrating the crucial role of AVI convergence.

A few works tried to understand AVI scheme and TD-learning in a unified perspective.~\cite{guo2022convex} proposed a convex program test approach for value iteration and TD-learning but requires different test for each algorithm.~\cite{wu2025unifying} provided an understanding of TD-learning and AVI from the matrix splitting technique~\citep{berman1994nonnegative}. In contrast, our work focuses on Q-learning, which presents unique analytical challenges due to non-linearity of the max-operator which makes the standard TD-learning analysis techniques insufficient.

Pathological behaviors regarding solution of the PBE, e.g, the non-existence or multiplicity of solutions, which can lead to suboptimal policies has been well-known in the literature~\citep{de2000existence,bertsekas2011approximate,young2020understanding}. This becomes more complex when we use $\eps$-greedy policy.\footnote{See Appendix~\ref{app:sec:related-works} for more detail.}{~\cite{lu2018non} provided an example that for a certain regime of $\eps$, Q-learning can yield a sub-optimal policy compared to possible ones that can be represented by the linear feature while the optimal policy is not realizable.} Covering a different scenario, we provide an example that the number of solutions depends on the choice of \(\eps\), and depending on $\eps$, there is a solution of PBE induces optimal policy but to which Q-learning cannot converge.




\section{Preliminaries}
\subsection{Markov decision process (MDP)}

MDP consists of five tuples $(\gS,\gA,\gamma,\gP,r)$. $\gS:=[|\gS|]$ and $\gA:=[|\gA|]$, where $[n]:=\{1,2,\dots,n\}$ for $n\in\sN$, are finite state and action spaces, respectively. $\gamma\in(0,1)$ is the discount factor. $\gP:\gS\times\gA\to \Delta^{\gS}$ is the Markov kernel where $\Delta^{\gS}$ denotes a probability distribution over the set $\gS$. $r:\gS\times\gA\times\gS\to\R$ is the reward function, which we assume to be bounded. An agent at state $s\in\gS$ selects an action $a\sim\pi(\cdot\mid s)$ following a policy $\pi:\gS\to\Delta^{\gA}$. Then, transition occurs to next state $s^{\prime}\sim\gP(\cdot\mid s,a)$ and the agent receives reward $r(s,a,s^{\prime})$. The $Q$-function induced by policy $\pi$ is defined as $Q^{\pi}(s,a)=\E\left[ \sum^{\infty }_{k=0}\gamma^k r(S_k,A_k,S_{k+1})\mid (S_0,A_0)=(s,a), \pi \right]$, where $\{(S_k,A_k)\in\gS\times \gA\}_{k=0}^{\infty}$ are a sequence of random variables following the policy $\pi$. The goal is to find an optimal policy $\pi^*$ such that $\pi^*=\argmax_{\pi\in\Omega}\E\left[ \sum^{\infty }_{k=0}\gamma^k r(S_k,A_k,S_{k+1})\mid  \pi \right]$ where $\Omega$ is the set of all deterministic policies. We denote $Q^*$ as the optimal $Q$-function, which is the $Q$-function induced by the optimal policy $\pi^*$. The optimal $Q$-function satisfies the Bellman optimality equation : $\mR+\gamma \mP\mQ^*=\mQ^*$ where $\mR\in\R^{|\gS||\gA|}$ is a vector such that $[\mR]_{(s-1)|\gA|+a}=\E\left[ r(s,a,s^{\prime}) \mid (s,a)\right]$, $\mP\in\R^{|\gS||\gA|\times|\gS|}$ is transition matrix such that $[\mP]_{(s-1)|\gA|+a,s^{\prime}} = \gP(s^{\prime}\mid s,a)$, and $\mQ^*\in\R^{|\gS||\gA|}$ is a vector such that $[\mQ^*]_{(s-1)|\gA|+a}=Q^*(s,a)$, where for $\vv\in\R^n$ and $i\in[n]$, $[\vv]_i$ denotes $i$-th element of $\vv$, and $[\mA]_{i,j}$ for $\mA\in\R^{n\times m}$ denotes the element in the $i$-th row and $j$-th column of $\mA$.

\subsection{Linear function approximation of $Q$-function}

Consider a set of features $\{\vphi(s,a)\in\R^p\}_{(s,a)\in\gS\times\gA}$, where $p\in\sN$ is the feature dimension. We approximate the $Q$-function , $Q^{\pi}(s,a)\approx \vphi(s,a)^{\top}\vtheta$ where $\vtheta\in\R^p$ is the learnable parameter. The $Q$-function may not be exactly represented by the feature, therefore we consider the following projected version of Bellman optimality equation~\citep{sutton2008convergent}, which is motivated from $\min_{\vtheta\in\R^p}\frac{1}{2}\left\| \vy-\mPhi\vtheta \right\|^2_2$ where $\vy=\mPhi(\mPhi^{\top}\mD_{\nu_{\vtheta}}\mPhi)^{-1}\mPhi^{\top}\mD_{\nu_{\vtheta}}(\mR+\gamma \mP\mPi_{\pi_{\vtheta}}\mPhi\vtheta)$:
\begin{align}
 \mF(\vtheta,\pi_{\vtheta},\nu_{\vtheta}):=&  \mPhi^{\top}\mD_{\nu_{\vtheta}}\mR+ \mT(\vtheta,\pi_{\vtheta},\nu_{\vtheta})\vtheta =  \bm{0},\label{pboe}\\
     \mT(\vtheta,\pi_{\vtheta},\nu_{\vtheta}):=& \gamma\mPhi^{\top}\mD_{\nu_{\vtheta}}\mP\mPi_{\pi_{\vtheta}}\mPhi-\mPhi^{\top}\mD_{\nu_{\vtheta}}\mPhi. \label{def:T}
\end{align}
where the sampling distribution \(\nu_{\vtheta}\in \Delta^{\gA}\) and the target policy \(\pi_{\vtheta}:\gS \to \Delta^{\gA}\) are parameterized by $\vtheta\in\R^p$, the matrix \(\mPhi \in \mathbb{R}^{|\gS||\gA| \times p}\) has its \((s-1)|\gA| + a\)-th row corresponding to the vector \(\vphi(s,a)^{\top}\), and the matrix \(\mPi_{\pi} \in \mathbb{R}^{|\gS|\times |\gS||\gA|}\) has the \(s\)-th row vector given by \((\ve_s \otimes \vpi(s))^{\top}\), where \(\vpi(s) \in \mathbb{R}^{|\gA|}\) satisfies \([\vpi(s)]_a = \pi(a \mid s)\) and \(\ve_s\) is the unit vector with a value of one at the \( s \)-th position and zeros elsewhere . The diagonal matrix \(\mD_{\nu_{\vtheta}} \in \mathbb{R}^{|\gS||\gA|\times |\gS||\gA|}\) has $(s-1)|\gA|+a$-th diagonal entry as \(\nu_{\vtheta}(s,a)\).  $\nu_{\vtheta}$ can be set as the stationary distribution induced by Markov chain using a behavior policy \(\beta_{\vtheta}:\gS\to\Delta^{\gA}\), which we denote as $\mu_{\beta_{\vtheta}}$. We assume it to be unique and existent, which is standard in the literature~\citep{meyn2024projected,liu2025linear}:
\begin{assumption}\label{assumption:statinoary-distribution}
    Every element in the closure of $\{\mP\mPi_{\beta_{\vtheta}}: \vtheta\in\R^p\}$ induces an irreducible and aperiodic Markov chain. 
\end{assumption}

Note that $\nu_{\vtheta}$ in~(\ref{pboe}) can be also set as some arbitrary fixed probability distribution $d\in\Delta^{\gS\times\gA}$ such $d(s,a)>0$ for all $(s,a)\in\gS\times\gA$ when we can sample state action pair from a fixed distribution, for example using a experience replay buffer~\citep{lin1992self}.


Meanwhile, the solution to (\ref{pboe}) may not exist. To ensure the existence of a solution, we can add an additional term \(\eta\vtheta\) (for some positive real number \(\eta\)) to (\ref{pboe}), which can be interpreted as the regularized PBE (\ref{reg-pboe})~\citep{zhang2021breaking,lim2024regularized}.
\begin{align}
\mF_{\eta}(\vtheta,\pi_{\vtheta},\nu_{\vtheta}):= \mPhi^{\top}\mD_{\nu_{\vtheta}}\mR+\mT(\vtheta,\pi_{\vtheta},\nu_{\vtheta})\vtheta-\eta \vtheta =  \bm{0}.\label{reg-pboe}
\end{align}

\section{Projected Bellman Equation}\label{sec:pbe}

In this section, we discuss the existence and uniqueness of solution of PBE in~(\ref{pboe}). It is known that the solution of~(\ref{pboe}) might not exist or there might be multiple depending on the choice of behavior policy and target policy~\citep{de2000existence,bertsekas2011approximate}. Section~\ref{sec:snrdd-existence} considers a condition using SNRDD and Section~\ref{sec:avi-existence} provides a condition motivated from the AVI algorithm. The relationship between these two conditions is thoroughly examined in Section~\ref{sec:discussion-snrdd-avi}.

\subsection{SNRDD guarantees existence and uniqueness of solution to~(\ref{pboe})}\label{sec:snrdd-existence}

Let us introduce a condition that guarantees the existence and uniqueness of the solution of~(\ref{pboe}). The key concept we leverage is the strictly negatively row dominating diagonal (SNRDD) condition:
\begin{definition}[~\cite{molchanov1989criteria}]\label{def:snrdd}
    A matrix $\mA\in\R^{n\times n}$ is said to have strictly negatively row dominating diagonal if  $S_i(\mA):=[\mA]_{i,i}+\sum_{j\in[n]\setminus\{i\} }[\mA]_{i,j}<0$ for all $i\in[n]$.
\end{definition}
For simplicity, we will call a matrix $\mA$ is SNRDD if it satisfies Definition~\ref{def:snrdd}. The above condition has been widely used in determining the stability of a dynamical system~\citep{molchanov1989criteria} or analysis of fixed point problem~\citep{davydov2024non}, which is summarized in Appendix Section~\ref{sec:fixed-point}. We explore the solution of PBE with this assumption and consider various behavior and target policy scenarios. Now, let us consider a parameterized form of SNRDD, for $\mM_{\vtheta}\in\R^{p\times p}$, a matrix dependent on $\vtheta$, and for some set $\gD\subseteq\R^p$:
\begin{align}
\sup_{\vtheta\in\gD}\max_{i\in[p]}S_i(\mM_{\vtheta})<0 . \label{cond:snrdd}
\end{align}
where $S_i$ is defined in Definition~\ref{def:snrdd}, and we call the above inequality as \textit{condition~(\ref{cond:snrdd}) with $(\gD,\mM_{\vtheta})$}.  

Depending on the choice of behavior and target policy, the existence of solution to PBE differs. A policy $\pi_{\vtheta}$ is said to be continuous if it is continuous with respect to $\vtheta$, and Lipschitz if $| \pi_{\vtheta}(a\mid s) - \pi_{\tilde{\vtheta}}(a\mid s) |\leq L\| \vtheta-\tilde{\vtheta} \|$ for some norm $\left\|\cdot\right\|$ and a positive real number $L$. Typical examples of Lipschitz policies are the greedy policy and the \(\epsilon\)-softmax policy, as discussed in Appendix~\ref{sec:choice-of-policy}. 

\begin{theorem}\label{prop:pbe-existence:1}
    \begin{enumerate}[wide, labelwidth=!, labelindent=0pt,itemsep=-0.5mm]
        \item Assume that both the behavior policy, $\beta_{\vtheta}$, and the target policy, $\pi_{\vtheta}$, are continuous. Suppose the parameterized SNRDD condition in~(\ref{cond:snrdd}) holds with $(\R^p,\mT(\vtheta,\pi_{\vtheta},\mu_{\beta_{\vtheta}}))$, where the later  defined in~(\ref{def:T}). Then, a solution of $\mF(\vtheta,\pi_{\vtheta},\mu_{\beta_{\vtheta}})=\bm{0}$ defined in~(\ref{pboe}) exists.
        \item Consider a fixed behavior policy, i.e., $\beta_{\vtheta}=\beta$ for some $\beta:\gS\to\Delta^{\gA}$, and a Lipschitz policy $\pi_{\vtheta}$. Suppose the condition in~(\ref{cond:snrdd}) holds with $(\gD_{F_{\mathrm{linear}}},\mT(\vtheta,\pi_{\vtheta},\mu_{\beta}))$ where $\gD_{F_{\mathrm{linear}}}$ is the set of all differentiable points of $\mF(\vtheta,\pi_{\vtheta},\mu_{\beta})$. Then, a solution of $\mF(\vtheta,\pi_{\vtheta},\mu_{\beta})=\bm{0}$ exists and is unique.
    \end{enumerate}
\end{theorem}
The proof, given in Appendix~\ref{app:prop:pbe-existence:1},  uses standard methods of fixed point theory~\citep{brouwer1911,banach1922operations}. 

\begin{remark}
~\cite{de2000existence} proved the existence of the solution when the behavior and target policy are identical (the on-policy case), and they are continuous. In contrast, we allow scenarios under different behavior and target policy, i.e., the off-policy case.~\cite{meyn2024projected} proved that using a particular type of $\epsilon$-softmax policy, so-called $(\eps,\kappa_0)$-tamed Gibbs policy (detailed in Appendix~\ref{sec:choice-of-policy}), ensures the existence of a solution of PBE. This covers different scenario from ours as using a $(\eps,\kappa_0)$-tamed Gibbs policy does not necessarily imply SNRDD condition.
\end{remark}

\begin{remark}
Given a Lipschitz target policy $\pi_{\vtheta}$ and fixed behavior policy $\beta$, $\mF(\vtheta,\pi_{\vtheta},\beta)$ is a Lipschitz function, which is differentiable almost everywhere by Rademacher's theorem~\citep{evans2018measure}.
\end{remark}
\begin{remark}
    The condition in~(\ref{cond:snrdd}) holds with $(\R^p,\mT(\vtheta,\pi_{\vtheta},\mu_{\beta_{\vtheta}}))$ when $\mPhi=\mI$ where $\mI$ is a $|\gS||\gA|\times |\gS||\gA|$ identity matrix, and behavior policy satisfies the condition $\inf_{\vtheta\in\R^p}\min_{(s,a)\in\gS\times\gA}\mu_{\beta_{\vtheta}}(s,a)>0$. This corresponds to the tabular setup of PBE, 
\end{remark}
   Considering the solution of PBE, as the feature dimension \( p \) increases, it becomes more challenging to satisfy condition (\ref{cond:snrdd}) due to the growing column size. One simple way to address this issue is to consider a matrix with additional scaled identity matrix, i.e., $\mT(\vtheta,\pi_{\vtheta},\mu_{\beta_{\vtheta}})-\eta\mI$ for $\eta>0$. This yields the regularized version of PBE given in~(\ref{reg-pboe}), and the same arguments in Theorem~\ref{prop:pbe-existence:1} hold for the solution to~(\ref{reg-pboe}). The SNRDD assumption can be satisfied with the following choice of $\eta$:
\begin{lemma}
    If $\eta> \sup_{\vtheta\in\R^p}\max_{i\in[p]} S_i(\mT(\vtheta,\pi_{\vtheta},\mu_{\beta_{\vtheta}}))$,~(\ref{cond:snrdd}) holds with $(\R^p,\mT(\vtheta,\pi_{\vtheta},\mu_{\beta_{\vtheta}})-\eta\mI)$.
\end{lemma}
\begin{remark}
When feature scaling is used, $\left\|\vphi(s,a)\right\|_{\infty}<1/\sqrt{p}$ for all $(s,a)\in\gS\times\gA$, then $\eta>3$ is sufficient to meet the above condition. The proof is given in Lemma~\ref{lem:eta>3->SNRDD} in Appendix~\ref{sec:omitted-proofs}.  
\end{remark}

\begin{remark}
    The SNRDD condition was also considered in~\citep{lim2024regularized} but only in terms of convergence of regularized Q-learning but not existence of solution, and it requires additional assumptions including positiveness and orthogonality on the feature matrix. In Section~\ref{sec:q-learning}, we show that only SNRDD condition is required for proving the convergence of regularized Q-learning.
\end{remark}

\begin{remark}
    For~(\ref{reg-pboe}), when $\mPhi=\mI$, then $\eta>0$ implies using a smaller discount factor, $\gamma$~\citep{chen2023target}. Nonetheless, the interpretation is more complex when $\mPhi\neq\mI$, and algorithms to sovle~(\ref{reg-pboe}) has been widely used in practice~\citep{farebrother2018generalization,cobbe2019quantifying}.
\end{remark}

\subsection{Condition motivated from AVI for existence and uniqueness of solution of~(\ref{pboe})}\label{sec:avi-existence}

Meanwhile, let us investigate another sufficient condition to guarantee the existence of solution of PBE in~(\ref{pboe}), which is motivated from the AVI algorithm. We can re-write~(\ref{pboe}) as
\begin{align}
\vtheta = (\mPhi^{\top}\mD_{\mu_{\beta_{\vtheta}}}\mPhi)^{-1}(\gamma\mPhi^{\top}\mD_{\mu_{\beta_{\vtheta}}}\mP\mPi_{\pi_{\vtheta}}\mPhi\vtheta+\mPhi^{\top}\mD_{\mu_{\beta_{\vtheta}}}\mR ) \label{pbe:2}
\end{align}
assuming invertibility of $\mPhi^{\top}\mD_{\mu_{\beta_{\vtheta}}}\mPhi$. Therefore, a closely related condition to guarantee the existence and uniqueness of the solution to~(\ref{pboe}) is that for \(\gD \subseteq \mathbb{R}^p\), a set to be defined further, one of the following two conditions hold:
\begin{numcases}{}
      \sup_{\vtheta\in\gD}\gamma \| \mPhi (\mPhi^{\top}\mD_{\mu_{\beta_{\mPhi\vtheta}}}\mPhi)^{-1}\mPhi^{\top}\mD_{\mu_{\beta_{\mPhi\vtheta}}}\mP\mPi_{\pi_{\mPhi\vtheta}} \|_{\infty} <1,  \label{suff-cond:avi-convrge:1}\\
         \sup_{\vtheta\in\gD} \gamma  \|  (\mPhi^{\top}\mD_{\mu_{\beta_{\vtheta}}}\mPhi)^{-1} \mPhi^{\top}\mD_{\mu_{\beta_{\vtheta}}}\mP\mPi_{\pi_{\vtheta}}\mPhi \|_{\infty} <1.\label{suff-cond:avi-convrge:2}
\end{numcases}
Note that the policies in~(\ref{suff-cond:avi-convrge:1}) are dependent on $\mPhi\vtheta$. As in Section~\ref{sec:pbe}, the following results can be derived using standard fixed point theory arguments, of which the proof deferred to Appendix~\ref{app:prop:pbe-existence:2}. 

\begin{theorem}\label{prop:pbe-existence:2}
    \begin{enumerate}[wide, labelwidth=!, labelindent=0pt]
        \item Suppose $\beta_{\vtheta}$ and $\pi_{\vtheta}$ are continuous. Moreover, assume that either~(\ref{suff-cond:avi-convrge:1}) or~(\ref{suff-cond:avi-convrge:2}) holds with $\gD=\R^p$, and  $0<\inf_{\vtheta\in\gD}\lambda_{\min}(\mPhi^{\top}\mD_{\mu_{\beta_{\vtheta}}}\mPhi)$. Then, a solution of~(\ref{pboe}) exists.
        \item Suppose a fixed behavior policy $\beta_{\vtheta}$ is used and $\pi_{\vtheta}$ is Lipschitz. Moreover, assume that either~(\ref{suff-cond:avi-convrge:1}) or~(\ref{suff-cond:avi-convrge:2}) holds with $\gD$ being all the differentiable points of $\mF(\vtheta,\pi_{\vtheta},\beta-{\vtheta})$, and $0<\inf_{\vtheta\in\gD}\lambda_{\min}(\mPhi^{\top}\mD_{\mu_{\beta_{\vtheta}}}\mPhi)$. Then, a solution of~(\ref{pboe}) exists and is unique.
    \end{enumerate}
\end{theorem}

\begin{remark}
    One can replace the infinity norm with joint spectral radius, which is defined as, given a set of square matrices $\{\mA_i\in\R^{n\times n}\}_{i=1}^m,m\in\sN$, $\rho(\mA_1,\cdots,\mA_m)= \lim_{k\to\infty}\max_{\sigma\in \{1,2,\dots,m\}^k}\left\| \mA_{\sigma_k}\cdots\mA_{\sigma_2}\mA_{\sigma_1} \right\|^{1/k} $. If $\rho(\mA_1,\mA_2,\dots,\mA_m)<1$, there exists a norm $\left\|\cdot\right\|$ such that $\left\|\mA_i\right\|<1$ for all $i\in[m]$~\citep{rota1960note}. Therefore, we can replace the infinity norm with this common norm in~(\ref{suff-cond:avi-convrge:1}) or~(\ref{suff-cond:avi-convrge:2}). 

It is important to note that each matrix $\mA_i$ having a spectral radius smaller than one — the maximum absolute value of its eigenvalues — does not imply Theorem~\ref{prop:pbe-existence:2}. This is because it does not guarantee the existence of a common norm \(\left\|\cdot\right\|\) such that \(\left\|\mA_i\right\| < 1\)~\citep{jungers2009joint}.
\end{remark}

\begin{remark}
    It is challenging to ensure when~(\ref{suff-cond:avi-convrge:1}) or~(\ref{suff-cond:avi-convrge:2}) hold in practice. Alternatively, one may consider a form motivated from the regularized PBE in~(\ref{reg-pboe}) by replacing $(\mPhi^{\top}\mD_{\mu_{\beta_{\vtheta}}}\mPhi)^{-1}$ with $(\mPhi^{\top}\mD_{\mu_{\beta_{\vtheta}}}\mPhi+\eta\mI)^{-1}$, and ensure the solution of~(\ref{reg-pboe}).
    
    ~\cite{zhang2021breaking} showed the existence of a solution to~(\ref{reg-pboe}), regularized PBE, whereas extension of Theorem~\ref{prop:pbe-existence:2} with regularization can guarantee uniqueness.~\cite{lim2024regularized} showed the uniqueness of the solution but we sharpen the bound from $\gamma \| \mPhi (\mPhi^{\top}\mD_{\mu_{\beta_{\mPhi\vtheta}}}\mPhi+\eta\mI)^{-1}\mPhi\mD_{\mu_{\beta_{\mPhi\vtheta}}} \|_{\infty}<1$ to $\gamma \| \mPhi (\mPhi^{\top}\mD_{\mu_{\beta_{\mPhi\vtheta}}}\mPhi+\eta\mI)^{-1}\mPhi\mD_{\mu_{\beta_{\mPhi\vtheta}}}\mP\mPi_{\pi_{\mPhi\vtheta}} \|_{\infty}<1$ from~(\ref{suff-cond:avi-convrge:1}). This follows from the application of a version of mean value theorem in Lemma~\ref{lem:lebourg-mvt} in the Appendix~\ref{app:sec:fine-property-function}.
\end{remark}
\subsection{Discussion on the condition~(\ref{cond:snrdd}) and~(\ref{suff-cond:avi-convrge:2}) (SNRDD and condition motivated form AVI)}\label{sec:discussion-snrdd-avi}
Letting $\mM_{\vtheta}=\mT_{\mu_{\beta_{\vtheta}}}(\vtheta,\pi_{\vtheta},\mu_{\beta_{\vtheta}}) $ in~(\ref{cond:snrdd}),  we now examine when the conditions~(\ref{cond:snrdd}) and~(\ref{suff-cond:avi-convrge:2}) imply each other. While either condition guarantees the existence of a solution of PBE, they are closely tied to the convergence of Q-learning and the AVI, respectively, which we defer the discussion to Section~\ref{sec:avi}. Below, we present a result on the relationship between conditions~(\ref{cond:snrdd}) and~(\ref{suff-cond:avi-convrge:2}).

\begin{proposition}\label{prop:pbe-splitting-contraction}
    If $\inf_{\vtheta\in\gD}\lambda_{\min}(\mPhi^{\top}\mD_{\mu_{\beta_{\vtheta}}}\mPhi)>0$ for some $\gD\subseteq\R^p$, the following holds: 
    \begin{enumerate}[wide, labelwidth=!, labelindent=0pt, itemsep=-0.5mm]
   \item[1)] Suppose~(\ref{cond:snrdd}) holds with $(\gD,\mT_{\mu_{\beta_{\vtheta}}}(\vtheta,\pi_{\vtheta},\mu_{\beta_{\vtheta}})) $, and $\mPhi^{\top}\mD_{\mu_{\beta_{\vtheta}}}\mP\mPi_{\pi_{\vtheta}}\mPhi$ has non-negative diagonal elements for all $\vx\in\gD$. Then,~(\ref{suff-cond:avi-convrge:2}) holds with $\gD$.
        \item[2)]   Suppose~(\ref{suff-cond:avi-convrge:2}) holds for $\gD\subseteq\R^p$. Then,~(\ref{cond:snrdd}) holds with $(\gD,\mT_{\mu_{\beta_{\vtheta}}}(\vtheta,\pi_{\vtheta},\mu_{\beta_{\vtheta}}))$.    
    \end{enumerate}
\end{proposition}

The proof is given in Appendix~\ref{app:prop:pbe-splitting-contraction}. If $\mPhi^{\top}\mD_{\mu_{\beta_{\vtheta}}}\mPhi$ is a diagonal matrix, and diagonal elements of $\mPhi^{\top}\mD_{\mu_{\beta_{\vtheta}}}\mP\mPi_{\pi_{\vtheta}}\mPhi$ are non-negative, then the conditions~(\ref{cond:snrdd}) and~(\ref{suff-cond:avi-convrge:2}) are equivalent. The diagonal elements of  $\mPhi^{\top}\mD_{\mu_{\beta_{\vtheta}}}\mP\mPi_{\pi_{\vtheta}}\mPhi$ can be non-negative if each entry of $\mPhi$ has non-negative values. 

We note that condition (\ref{suff-cond:avi-convrge:1}) also guarantees solution existence, though its relationship with condition (\ref{cond:snrdd}) is difficult to characterize. As these conditions are linked to AVI and Q-learning convergence respectively, in Section~\ref{sec:avi}, we present an example where only one condition is met, causing only its corresponding algorithm to converge while the other fails. Moreover, it is not clear whether we can construct such example with condition~(\ref{suff-cond:avi-convrge:2}), which requires further research.

\begin{figure}
\centering
\begin{subfigure}[t]{0.3\textwidth}
    \centering
    \includegraphics[height=2.8cm]{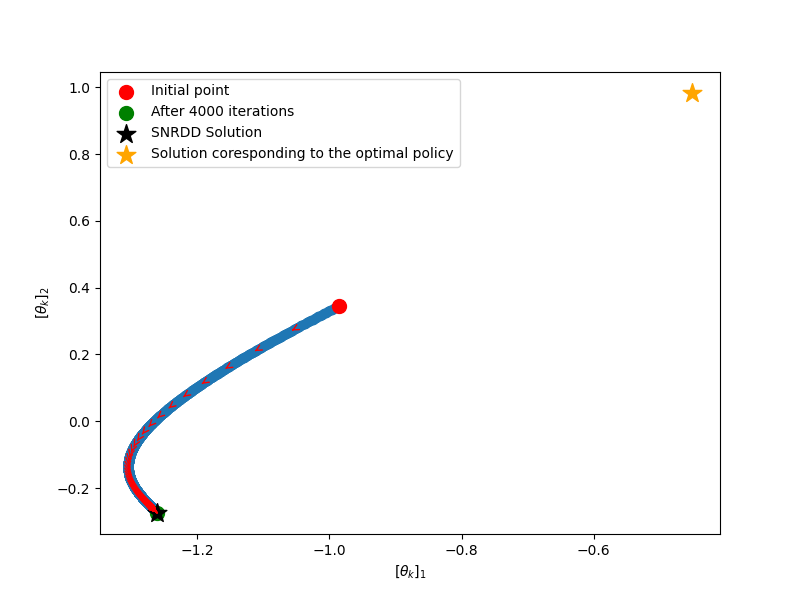}
\caption{Linear Q-learning converges to a point which induces a sub-optimal policy.}\label{fig:snrdd-sub-optimal-traj}
\end{subfigure}
\begin{subfigure}[t]{0.3\textwidth}
    \centering
    \includegraphics[height=2.8cm]{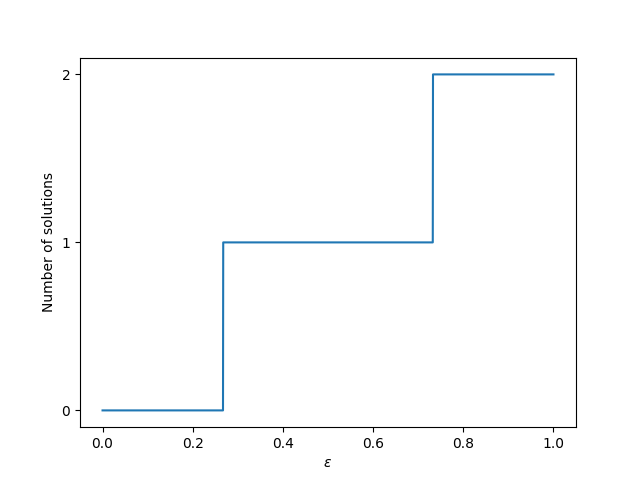}
\caption{If $\eps$ is small, there is no solution, and if $\eps$ is large, multiple solutions exist.}\label{fig:eps-solution-change-eps}
\end{subfigure}
\begin{subfigure}[t]{0.3\textwidth}
    \centering
    \includegraphics[height=2.8cm]{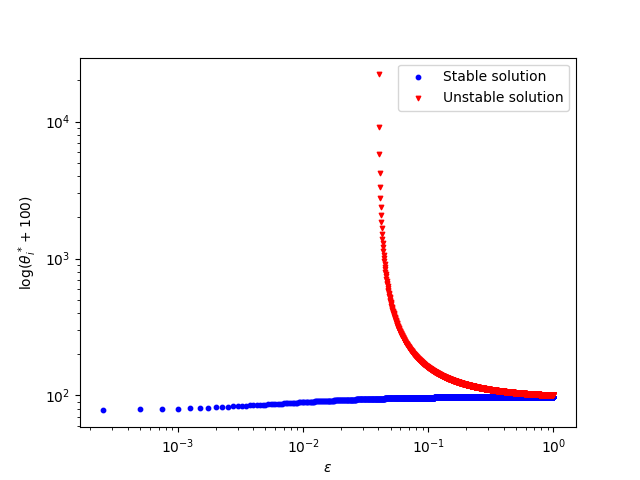}
\caption{Increasing $\eps$ adds an unstable solution. $r_1=-0.1$ and $r_2=-0.78$.}\label{fig:eps-unstable}
\end{subfigure}
\caption{The first and last two figures show Example~\ref{ex:snrdd:local-minima}  and~\ref{ex:eps-unstable} in Appendix~\ref{sec:mdp_examples}, respectively. In the last figure, stable and unstable refers to whether $\mT(\vtheta,\pi_{\vtheta},\beta_{\vtheta})$ is a Hurwitz matrix at each point.} 
\end{figure}

\section{Convergence of Q-learning}\label{sec:q-learning}
In this section, we briefly review the Q-learning algorithm, and prove its convergence by ordinary differential equation (ODE) analysis using the parameterized SNRDD condition in~(\ref{cond:snrdd}). We consider an i.i.d. sampling model from an arbitrary fixed distribution $d\in \Delta^{\gS\times\gA}$. An experience replay buffer~\citep{lin1992self} can mimic the i.i.d. sampling model and our analysis can be easily extended to the Markovian observation model using the arguments in~\citep{liu2024ode}. Upon observing $(s_k,a_k)\sim d(\cdot),\;s_k^{\prime}\sim\gP(\cdot\mid s_k,a_k)$ and $r_k:=r(s_k,a_k,s_k^{\prime})$ independently for every $k$-th iteration, the parameter of (regularized) Q-learning using step-size \footnote{This conditions is called the Robbins-Monro step-size condition~\citep{robbins1951stochastic}.}{$\alpha_k\in (0,1)$ satisfying $\sum_{k\in\sN}\alpha_k=\infty, \sum_{k\in\sN}\alpha_k^2<\infty$} is updated as follows:
\begin{align}
\vtheta _{k + 1} = \vtheta _k + \alpha _k \vphi(s_k,a_k)(r_k+\gamma \max_{a\in\gA}\vphi^{\top}(s_{k}^{\prime},a)\vtheta_k-\vphi(s_k,a_k)^{\top}\vtheta_k - \eta\vtheta_k), \quad \vtheta_0 \in \R^p \label{eq:q-learning}
\end{align}
\subsection{Stochastic approximation and ODE approach} Q-learning can be understood as a  stochastic approximation scheme~\citep{robbins1951stochastic}:
\begin{align}
    \vx_{k+1} = \vx_k + \alpha_k (\mA_{\sigma(\vx_k)}\vx_k+\vb+ \veps_{k+1}),\quad \vx_0\in\R^p \label{eq:sa-switcehd-system}
\end{align}
where $\sigma:\R^p\to\gM$ is a switching signal, $\gM:=\{1,2,\dots,|\gM|\}$ is the set of modes, $\vb\in\R^p$ is a constant vector and $\{\mA_{m} : m\in \gM\}$ are the subsystem matrices. $\veps_k$ is Martingale-difference sequence defined in Definition~\ref{def:mds} in the Appendix. The almost sure convergence of~(\ref{eq:sa-switcehd-system}) is closely related to its ODE counterpart:
 \begin{align}
    \dot{\vx}_t = \mA_{\sigma(\vx_t)}\vx_t+\vb,\quad \vx_0\in \R^p , t \geq 0\label{eq:ode-switched-system}
\end{align}
where $\frac{d}{dt}\vx_t=\dot{\vx}_t$. Loosely speaking, the asymptotic behavior of $\vx_k$ in~(\ref{eq:sa-switcehd-system}) is governed by its corresponding ODE if it admits a globally asymptotically stable equilibrium point. An equilibrium point is a vector $\vx^*\in\R^p$ that satisfies $ \mA_{\sigma(\vx^*)}\vx^*+\vb=\bm{0}$ and global asymptotic stability means that the solutions $\vx_t$ converge to $\vx^*$ regardless of the initial condition $\vx_0$. A detailed argument is given by Borkar and Meyn Theorem~\citep{borkar2000ode} provided in Appendix~\ref{sec:sa}. A key concept in verifying a globally asymptotically stable equilibrium point is the so-called one-sided Lipschitznes:

\begin{definition}[One-sided Lipschitz, Definition 3.2 in~\cite{FB-CTDS}]\label{def:one-sided-Lipschitz}
For $\gD\subseteq\R^p$, if $\vf:\gD\to\R^p$ satisfies the following it is called one-sided Lipschitz with constant $b$: $[\vf(\vx)-\vf(\vy)]_i [\vx-\vy]_i \leq  b \left\| \vx-\vy \right\|^2_{\infty}$ where $i\in \gI_{\infty}(\vx-\vy):=\left\{ j\in [p] :  |[\vx-\vy]_j|=\left\| \vx-\vy\right\|_{\infty} \right\}$ and $\vx,\vy\in\R^p$.
\end{definition}

If \( \vf \) is one-sided Lipschitz with a negative constant, then every pair of trajectories of (\ref{eq:ode-switched-system}) are contracting, i.e., for solutions \( \vx_t \) and \( \vy_t \) with different initial conditions \( \vx_0 \) and \( \vy_0 \), we have \( \|\vx_t - \vy_t\|_{\infty} \to 0 \). This is known as contraction theory~\citep{lohmiller1998contraction}, and if a unique equilibrium exists, all trajectories converge to it.

\begin{lemma}[Theorem 3.9 in~\cite{FB-CTDS}]\label{lem:ode-contraction}
    Suppose Definition~\ref{def:one-sided-Lipschitz} holds for $\vf(\vx):=\mA_{\sigma(\vx)}\vx+\vb$ for $\R^p$ with some $c<0$ and $\vf$ is a Lipschitz function. Then, there exists a unique $\vx^*\in\R^p$ such that $\mA_{\sigma(\vx^*)}\vx^*+\vb=\bm{0}$ which is globally asymptotically stable.
\end{lemma}
 In fact, for a locally Lipschtiz function $\vf:\R^p\to\R^p$, Definition~\ref{def:snrdd} holding for $\nabla \vf(\vx)$--the gradient at differentiable point of $\vf$--at all such points is equivalent to the one-sided Lipschitz condition~\citep{davydov2024non} (see Lemma~\ref{lem:one-sided-snrdd-equivalence} in Appendix~\ref{app:sec:fine-property-function}).

\subsection{Analysis of Q-learning algorithms} 

Now, using the ODE arguments introduced in the previous section, we prove that ODE counterparts of a family of Q-learning algorithms admits a globally asymptotically stable equilibrium point. Let us consider the following ODE counterpart of the Q-learning algorithm in~(\ref{eq:q-learning}):
\begin{align*}
   \dot{\vtheta}_t =  \mPhi^{\top}\mD_d\mR+\gamma\mPhi^{\top}\mD_d\mP\mPi_{\pi^g_{\vtheta_t}}\mPhi\vtheta_t-(\mPhi^{\top}\mD_d\mPhi+\eta\mI)\vtheta_t,\quad \vtheta_0\in\R^p,\; t\geq 0,
\end{align*}
where $\pi^g_{\vtheta}$ denotes the greedy policy over $\mPhi\vtheta$, i.e., $\pi^g_{\vtheta}(s)=\argmax_{a\in\gA}\vphi(s,a)^{\top}\vtheta $, and a fixed tie-breaking rule is applied when it is not a singleton set. When \( \eta = 0 \), it coincides with the update rule of linear Q-learning, and if additionally \( \mPhi = \mI \), the algorithm reduces to asynchronous tabular Q-learning. If \( \eta > 0 \), the algorithm becomes regularized Q-learning. For each case, we can verify that the one-sided Lipschitz condition in Definition~\ref{def:one-sided-Lipschitz} holds under the SNRDD condition in~(\ref{cond:snrdd}) (which is necessary and sufficient condition for a locally Lipschitz map~\citep{davydov2024non}):

\begin{lemma}\label{lem:q-learning-osl} The following holds depending on the choice of $\mPhi$ and $\eta$:
\begin{enumerate}[wide, labelwidth=!, labelindent=0pt, itemsep=-0.5mm]
    \item[1)]   Let $\eta=0$ and $\mPhi=\mI$. For $\mQ\in\R^{|\gS||\gA|}$, let $\mF_{\mathrm{AsyncQ}}(\mQ)=\mF(\mQ,\pi^g_{\mQ},d)$. Then, $\mF_{\mathrm{AsyncQ}}(\mQ)$ is one-sided Lipschitz with constant $(\gamma-1)d_{\min}$ where $d_{\min}:=\min_{(s,a)\in\gS\times\gA}d(s,a)$.
    \item[2)]Let $\eta=0$ and suppose~(\ref{cond:snrdd}) holds with $(\gD_{F_{\mathrm{linear}}},\mT(\vtheta,\pi^g_{\vtheta},d) )$ where $\gD_{F_{\mathrm{linear}}}$ is the set of differentiable points of $\mF(\vtheta,\pi^g_{\vtheta},d)$. Let $\mF_{\mathrm{linear}}(\vtheta):=\mF(\vtheta,\pi^g_{\vtheta},d)$. Then, $\mF_{\mathrm{linear}}(\vtheta)$ is one-sided Lipschitz with constant $a_{\min}:=\sup_{\vtheta\in\gD_{F_{\mathrm{linear}}}}\max_{i\in [p]}S_i(\mT(\vtheta,\pi^g_{\vtheta},d))$ which is defined in Definition~\ref{def:snrdd}.
    \item[3)]Let $\eta> \sup_{\vtheta\in\gD_{F_{\mathrm{linear}}}}\max_{i\in [p]}S_i(\mT(\vtheta,\pi^g_{\vtheta}),d)$. Define $\mF_{\mathrm{Reg}}(\vtheta):=F_{\eta}(\vtheta,\pi^g_{\vtheta},d)$. Then, $\mF_{\mathrm{Reg}}(\vtheta)$ is one-sided Lipschitz with constant $-\eta+a_{\min}$ with respect to $\vtheta$.
\end{enumerate}
\end{lemma}

The proof is given in Appendix~\ref{app:lem:q-learning-osl}. The SNRDD condition ensures the uniqueness and existence of a solution, which corresponds to the globally asymptotically stable equilibrium point of  the ODE counterpart of each Q-learning algorithms by Lemma~\ref{lem:ode-contraction}. This yields the following result of which the proof is given in Appendix~\ref{app:sec:prop:q-learning-convergence}:

\begin{proposition}\label{prop:q-learning-convergence}
\begin{enumerate}[wide, labelwidth=!, labelindent=0pt, itemsep=-0.5mm]
    \item[1)] (Asynchronous tabular Q-learning) Let $\mPhi=\mI$ and $\eta=0$ in~(\ref{eq:q-learning}). Then, $\vtheta_k$ in~(\ref{eq:q-learning}) converges to a solution of $\mF(\vtheta,\pi^g_{\vtheta},d)=\bm{0}$ which is unique, with probability one.
    \item[2)] (Linear Q-learning) Let $\eta=0$ in~(\ref{eq:q-learning}). Suppose the parameterized SNRDD condition~(\ref{cond:snrdd}) holds with $(\gD_{F_{\mathrm{linear}}},\mT(\vtheta,\pi^g_{\vtheta},d))$ where $\gD_{F_{\mathrm{linear}}}$ is the set of differentiable points of $\mF(\vtheta,\pi^g_{\vtheta},d)$. Then, $\vtheta_k$ in~(\ref{eq:q-learning}) converges to the unique solution of $\mF(\vtheta,\pi^g_{\vtheta},d)=\bm{0}$ with probability one.
    \item[3)] (Regularized Q-learning) Let $\eta$ satisfy the condition in~(\ref{cond:snrdd}) with $(\gD_{F_{\mathrm{linear}}},\mT(\vtheta,\pi^g_{\vtheta},d)-\eta\mI)$.  Then, $\vtheta_k$ in~(\ref{eq:q-learning}) converges to the unique solution of $\mF_{\eta}(\vtheta,\pi^g_{\vtheta},d)=\bm{0}$ with probability one.
\end{enumerate}
\end{proposition}

\begin{remark}
To author's best knowledge, SNRDD assumption is the first assumption that can be both applied to prove convergence of linear and tabular Q-learning. As for regularized Q-learning, we relax the assumptions on positiveness and orthogonality of feature matrix~\citep{lim2024regularized}. 
\end{remark}

\begin{remark}[Convergence to sub-optimal policy]
    Even though Q-learning can converge to an unique point, there is no guarantee that this point induces the optimal policy. Moreover, suppose there exist multiple solutions of PBE. The matrix in~(\ref{cond:snrdd}) corresponding to one of the solution can be SNRDD while yielding a sub-optimal policy compared to the others. Then, the Q-learning algorithm may converge to this solution. A simple example is given in Example~\ref{ex:snrdd:local-minima} in the Appendix~\ref{sec:mdp_examples}, and its trajectories are shown in Figure~(\ref{fig:snrdd-sub-optimal-traj}). This complements the observation by~\cite{gopalan2024should}, which empirically showed that linear Q-learning can converge to the worst policy. 
\end{remark}


\section{Approximate value iteration and Q-learning}\label{sec:avi}
In this section, we analyze the convergence of the AVI algorithm and present examples where Q-learning converges while AVI does not, and vice versa. The convergence of AVI is known to play key role in algorithm with target network updates~\citep{lee2020periodic, chen2023target}. Nonetheless, its relation with Q-learning has not been not well understood.

\begin{figure}[t]
    \centering
\begin{subfigure}[t]{0.24\textwidth}
    \centering
    \includegraphics[height=3cm]{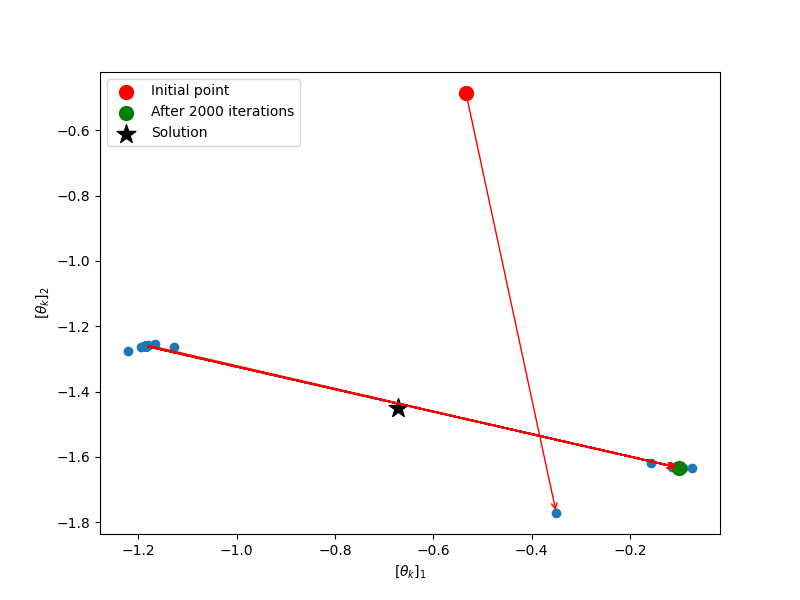}
\caption{Oscillation of AVI.}
\end{subfigure}
\hfill
\begin{subfigure}[t]{0.24\textwidth}
    \centering
    \includegraphics[height=3cm]{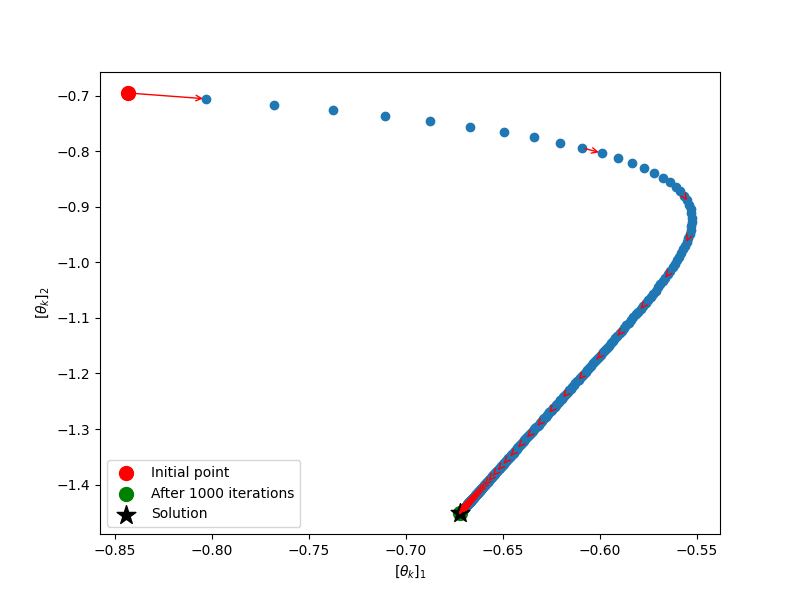}
\caption{Convergence of linear Q-learning.}
\end{subfigure}
\hfill
\begin{subfigure}[t]{0.24\textwidth}
    \centering
    \includegraphics[height=3cm]{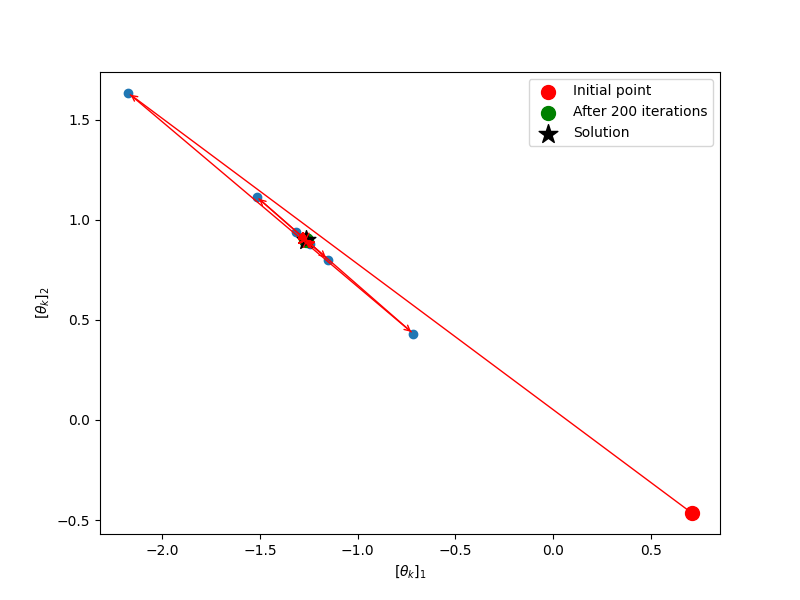}
\caption{Convergence of AVI.}
\end{subfigure}
\hfill
\begin{subfigure}[t]{0.24\textwidth}
    \centering
    \includegraphics[height=3cm]{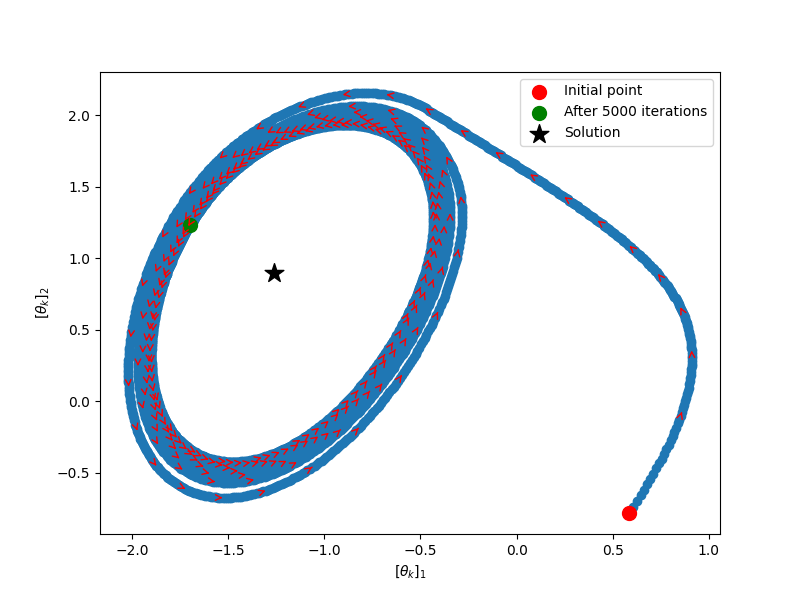}
\caption{Oscillation of linear Q-learning.}
\end{subfigure}
\caption{ The first two  and last two figures show experimental results on Example~\ref{ex:all-snrdd-but-not-avi-convergence} and~\ref{ex:all-avi-but-hurwitz}, respectively. For reproducibility, the experiments are done with an expected update version of Q-learning provided in Algorithm~\ref{algo:deterministic-q} in Appendix~\ref{sec:algo}.}\label{fig:avi-q-learning-convergence}
\end{figure}

An iterative method to solve~(\ref{pbe:2}), the so-called AVI algorithm~\citep{de2000existence}, is
\begin{align}
    \vtheta_{k+1} = (\mPhi^{\top}\mD_d\mPhi)^{-1}\mPhi^{\top}\mD_d(\mR+\gamma\mP\mPi_{\pi^g_{\vtheta_k}}\mPhi\vtheta_k), \quad \vtheta_0\in\R^p, k\in\sN,\label{avi} 
\end{align}
One can easily check that the condition in~(\ref{suff-cond:avi-convrge:1}) or~(\ref{suff-cond:avi-convrge:2}) ensures the convergence of~(\ref{avi}), which is given in Lemma~\ref{lem:convergence-of-avi} in the Appendix. Now, our focus is on the relation between the convergence of AVI and Q-learning. Proposition~\ref{prop:pbe-splitting-contraction} states a condition when both algorithms converge. Our interest is in the case when one algorithm converges while the other diverges. In particular, we consider the case when the condition~(\ref{cond:snrdd}) is met but the spectral radius of the matrix in~(\ref{suff-cond:avi-convrge:1}) at the solution is larger than one, i.e., the Q-learning converges but AVI does not. \footnote{To be precise, the expected version of Q-learning does not converge to a solution when $F(\vtheta)$ in~(\ref{pboe}) is differentiable and $\mT(\vtheta,\pi^g_{\vtheta},\beta_{\vtheta})$ is not a Hurwitz matrix at the corresponding point. The stochastic counterpart closely follows the behavior of expected update version.}{Likewise, we provide the opposite direction, Q-learning converges but AVI does not, by considering the case when condition~(\ref{suff-cond:avi-convrge:1}) is met but the matrix in~(\ref{cond:snrdd}) at the solution is not a Hurwitz matrix.} The examples are provided in Example~\ref{ex:all-snrdd-but-not-avi-convergence} and~\ref{ex:all-avi-but-hurwitz} in the Appendix~\ref{sec:mdp_examples}, and the experimental results are plotted in Figure~(\ref{fig:avi-q-learning-convergence}).

\begin{remark}
    If $|\gA|=1$, which is the case of TD-learning, the spectral radius of the matrix in~(\ref{suff-cond:avi-convrge:1}) being smaller than one is sufficient to guarantee the convergence of AVI. Moreover, the convergence of linear Q-learning can be checked whether the matrix $T(\vtheta,\pi^g_{\vtheta},d)$ is Hurwitz, i.e., the real part of the eigenvalues are all negative. Using these conditions,~\cite{wu2025unifying} provided an example that TD-learning converges but AVI does not, and vice-versa. In constrast, we provide examples for the case when $|\gA| \geq 2$. The spectral radius condition and Hurwitz conditions do not imply convergence of AVI and Q-learning, respectively. Moreover, SNRDD matrix is a Hurwitz matrix but the reverse does not necessarily hold, which is provided in Lemma~\ref{lem:snrdd-then-hurwitz} in the Appendix. Therefore, our example covers different scenarios from the example in~\cite{wu2025unifying}.
\end{remark}

\section{Pathological behavior using $\eps$-greedy behavior policy}\label{sec:discontinuous}
In this section, we examine the case when $\epsilon$-greedy policy is used, which was not addressed in the previous analysis. The first example illustrates a problem arising from this discontinuity, while the second example highlights a specific phenomenon resulting from the use of the $\epsilon$-greedy policy.

\textbf{Change of number of solutions:}  In this example, there is a critical value for $\epsilon$, at which the number of solutions to equation~(\ref{pboe}) changes. $\eps$-greedy policy is used for both behavior and target policy. Consider an MDP with $|\gS|=1,|\gA|=2,p=1,\gamma=0.99$, and $r(1,1,1)=0.5$ and $r(1,2,1)=0.48$, illustrated in Example~\ref{ex:eps-greedy-change-of-solution} in Appendix~\ref{app:sec:eps}. Depending on $\epsilon$, there are three distinct regions: no solution, unique solution, and multiple solutions (Figure~\ref{fig:eps-solution-change-eps}). A critical value separates these regimes, with the number of solutions changing when this threshold is crossed. In particular, as the SNRDD condition is met, the non-existence of the solution is due to using a discontinuous policy.

~\cite{bertsekas2011approximate,young2020understanding} provided examples that the number of solutions changes depending on the value of transition probability or reward. Our example differs as the change is determined by the value of $\eps$, which reflects the degree of exploration.

\textbf{Emergence of solution that yields optimal policy but to which Q-learning cannot converge:} We provide an example showing that increasing $\eps$ introduces a solution that induces optimal policy but to which Q-learning cannot converge,  which is illustrated in Figure~(\ref{fig:eps-unstable}). Consider an MDP with $|\gS|=1, |\gA|=2$, and $\vphi(1,1)=x, \vphi(1,2)=y$, and the behavior  and target policy are $\eps$-greedy and greedy policy, respectively. The reward $r(1,1,1)=r_1$ and $r(1,2,1)=r_2$ are negative constants, as illustrated in Example~\ref{ex:eps-unstable} in Appendix~\ref{app:sec:eps}. There are two possible deterministic policies, say $\pi_1$ and $\pi_2$. Depending on the choice of $r_1$ and $r_2$ ($r_1<r_2$ or $r_2<r_2$), the optimal policy can be either $\pi_1$ or $\pi_2$. When $r_1=-0.1<r_2=-0.78$, for $\eps<\eps^*\approx 0.1$, there exists a unique solution to PBE. This induces a sub-optimal policy, and Q-learning converges to this solution. For \(\epsilon > \epsilon^*\), there exist two solutions: one leading to a sub-optimal policy and the other to the optimal one. However, Q-learning cannot converge to the optimal solution because $F(\vtheta)$ in~(\ref{pboe}) is differentiable and \(\mT(\vtheta,\pi_{\vtheta},\beta_{\vtheta})\) is not Hurwitz at the corresponding point.  

~\cite{young2020understanding} provided an example that Q-learning can converge to a point that induces sub-optimal policy depending on the ordering of the reward (but not dependent on $\eps$).~\cite{lu2018non} showed that for a certain regime of $\eps$, Q-learning can yield a sub-optimal policy compared to possible policies that can be represented by the linear feature while the optimal one is not realizable (detailed in Appendix~\ref{app:sec:related-works}). Our example shows a different scenario that we can tune $\eps$ to ensure a solution of PBE that induces optimal policy, but Q-learning cannot converge to this solution.

\section{Conclusion}
In this paper, we have studied PBE through the lens of SNRDD assumption and condition motivated from the AVI scheme. We studied the relation of convergence of Q-learning and AVI. Moreover, to extend the understanding of solution to PBE, we provided examples showing peculiar phenomena when using $\eps$-greedy policy. Future studies would include considering non-linear function approximation.

\bibliographystyle{apalike}
\bibliography{biblio}

\section{Notations and organization}

\textbf{Notations:} $\R$ : set of real numbers; $\R_+$: set of non-negative real numbers; $\R^d$ : set of real-valued $d$-dimensional vectors, $\R^{m\times n}$ : set of real-valued $m\times n$-dimensional matrices; $\sC$ : set of complex numbers ; $[n]$ for $n\in\sN$ : $\{1,2,\dots, n\}$; $[\vv]_i$ for $\vv\in\R^n$ and $i\in[n]$ : $i$-th element of $\vv$; $[\mA]_{i,j}$ for $\mA\in\R^{n\times m}$: the element in the $i$-th row and $j$-th column of the matrix $\mA$; $\left\|\mA\right\|_{\infty}$ for $\mA\in\R^{m\times n}$: infinity norm of matrix $\mA$, i.e., $\max_{1\leq i \leq m} \sum_{j=1}^n|[\mA]_{i,j}|$; $\Delta^{\gD}$ for some set $\gD$ : a probability distribution over the set $\gD$; $\lambda_{\min}(\mA)$ for $\mA\in\R^{n\times n}:$ minimum eigenvalue of $\mA$;

\textbf{Organization:} 
In Section~\ref{sec:pbe}, conditions for existence of a solution of PBE is discussed. Section~\ref{sec:q-learning} and~\ref{sec:avi} provides convergence result for Q-learning and AVI, respectively. Lastly in Section~\ref{sec:discontinuous}, we discuss the properties of the solution of PBE when an $\eps$-greedy policy is adopted.

\section{Fixed point problem}\label{sec:fixed-point}

In this section, we present an analysis of existence and uniqueness for a specific equation. The results will be applied to the study of the solution of PBE in Section~\ref{sec:pbe}. Our goal is to solve the following equation:
\begin{align}
   h(\vx):=  \mA_{\vx}\vx + \vb_{\vx}=\bm{0}. \label{eq:pwa}
\end{align}
where $\mA_{\vx}\in\R^{p\times p}$ and  $\vb_{\vx}\in \R^p$ are a matrix and a vector that depend on $\vx\in\R^p$, respectively. When there are only finitely many possible choices of $\mA_{\vx}$ and $\vb_{\vx}$, it is called a switched affine system or a piecewise affine system. 

Finding a solution of~(\ref{eq:pwa}) can be re-casted into a fixed point problem: $\vx+\alpha h(\vx)=\vx$ for some $\alpha\in\R$. The study of fixed-point problems has been extensively explored in the literature, with foundational contributions from pioneering works such as~\cite{brouwer1911} and~\cite{banach1922operations}.

A common method for verifying the existence of a fixed point is to check if the matrix $\mA_{\vx}$ satisfies a specific condition. We focus on a matrix with a strictly negatively row-dominant diagonal introduced in Definition~\ref{def:snrdd}:
\begin{align}
   \sup_{\vx\in\gD}\max_{i\in[p]} S_i(\mA_{\vx})<0, \label{cond:snrdd-2}
\end{align}
where $S_i(\mA_{\vx})$ is defined in Definition~\ref{def:snrdd}, and $\gD\subset\R^p$ will be will be formally defined later. This concept has been widely used in the literature of fixed point problem as well as in various system analyses~\citep{molchanov1989criteria,davydov2024non}.

Now, let us present a simple result that follows from standard argument of Brouwer's fixed point theorem given in Lemma~\ref{lem:brouwer} in Appendix~\ref{app:prelim}: 
\begin{lemma}\label{lem:continuity-fixed-point}
Suppose~(\ref{cond:snrdd-2}) holds with $\gD=\R^p$, and $\sup_{\vx\in\R^p}\left\|\vb_{\vx}\right\|_{\infty}<\infty$. Furthermore, if the function \( h \) in (\ref{eq:pwa}) is continuous, then, a solution of~(\ref{eq:pwa}) exists. 
\end{lemma}

\begin{proof}
For simplicity of the proof, denote $c:= \sup_{\vx\in\R^p} \max_{i\in[p]}S_i(\mA_{\vx})$, which is a negative constant. Consider the following map : $H(\vx)=\vx+\alpha h(\vx)$ for small enough $\alpha$ such that $0<\alpha< \frac{1}{\sup_{\vx\in\R^p}\left\| \mA_{\vx}\right\|_{\infty}}$. Then, we have,
\begin{align*}
    \left\| \mI+\alpha \mA_{\vx} \right\|_{\infty}=&\max_{1\leq i \leq p} | 1+\alpha[\mA_{\vx}]_{i,i}| + \alpha \sum_{j\in\{1,2,\dots,p\}\setminus \{i\}}| [\mA_{\vx}]_{i,j}|\\
    = & \max_{1\leq i \leq p} 1+\alpha\left([\mA_{\vx}]_{i,i}  + \sum_{j\in\{1,2,\dots,p\}\setminus \{i\}}| [\mA_{\vx}]_{i,j}|\right) \\
    \leq & 1 + c \alpha .
\end{align*}
The second equality follows from the fact that $0<1+\alpha[\mA_{\vx}]_{i,i}$ from the choice of $\alpha$. The last inequality follows from the definition of $S_i(\mA_{\vx})$ in Definition~\ref{def:snrdd}.

Then, for $\vx$ such that $\left\|\vx \right\|_{\infty}\leq \frac{\sup_{\vx\in\R^p}\left\|\vb_{\vx}\right\|_{\infty}}{|c|} $,

\begin{align*}
    \left\| H(\vx) \right\|_{\infty} \leq &  \left\|  (\mI+\alpha \mA_{\vx})\vx\right\|_{\infty}+\alpha\left\| \vb_{\vx}\right\|_{\infty}\\
    \leq &(1+ c\alpha) \left\| \vx\right\|_{\infty}+\alpha\left\|\vb_{\vx}\right\|_{\infty}\\
    \leq & \frac{\sup_{\vx\in\R^p}\left\|\vb_{\vx}\right\|_{\infty}}{|c|} .
\end{align*}
    Therefore, $H$ is a self-map. Moreover, $H$ is a continuous function from the assumption. Therefore, by Brouwer's fixed point theorem in Lemma~\ref{lem:brouwer} in the Appendix, the existence of a fixed point of the map $H$ follows. 
\end{proof}

\begin{remark}
    Without continuity, characterizing the existence of a solution becomes challenging. An example where condition~(\ref{cond:snrdd-2}) is satisfied, yet no solution exists or multiple solutions arise, is provided in Section~\ref{sec:discontinuous}.
\end{remark}

If we assume a slightly more contingent assumption that $h$ in~(\ref{eq:pwa}) is a locally Lipschitz map (where the definition is given in Definition~\ref{def:locally-Lipschitz} in Appendix~\ref{app:prelim}), then we can guarantee a stronger result, i.e., the uniqueness of the solution, which is a result from~\cite{davydov2024non}:

\begin{lemma}[Lemma 24 in~\cite{davydov2024non}]\label{lem:snrdd-uniqueness-existence}
    Suppose \(\vb_{\vx}\) is a constant vector and \(h\) is a locally Lipschitz mapping, with condition~(\ref{cond:snrdd-2}) holding at all differentiable points of \(h\). Then, there exists a unique point $\vx^*\in\R^p$ such that $\mA_{\vx^*}\vx^*+\vb_{\vx^*}=\bm{0}_p$ where $\bm{0}_p$ is a zero vector in $\R^p$.
\end{lemma}

\begin{remark}
The continuity of a function does not necessarily imply Lipschitzness. However, if a function is Lipschitz, it is necessarily continuous.
\end{remark}
\begin{remark}
    A locally Lipschitz function is differentiable almost everywhere by Rademacher's theorem in Lemma~\ref{lem:rademacher} in Appendix~\ref{app:prelim}.
\end{remark}





Now, let us focus on a slightly different condition to study the solution of~(\ref{eq:pwa}). For some $\mC_{\vx}\in\R^{p\times p}$ that is invertible, we can re-write the equation in~(\ref{eq:pwa}) by
\begin{align}
    \tilde{h}(\vx):=   \mC^{-1}_{\vx}(\mA_{\vx}+\mC_{\vx})\vx+\mC^{-1}_{\vx}\vb_{\vx}  - \vx=\bm{0}. \label{eq:tilde-h}
\end{align} 

When $\mA_{\vx}$, $\vb_{\vx}$, and $\mC_{\vx}$ are constant matrices and vectors, i.e., $\mA_{\vx} = \mA$, $\vb_{\vx} = \vb$, and $\mC_{\vx} = \mC$ for some $\mA,\mC\in\R^{p\times p},\vb\in\R^p$, the system simplifies to a linear form. In this scenario, the reformulation in~(\ref{eq:tilde-h}) is widely recognized as matrix splitting~\citep{berman1994nonnegative}, a method extensively studied for analyzing the convergence of linear systems such as 
\begin{align*}
  \vx_{k+1} = \mC^{-1}(\mA + \mC)\vx_k,\quad \vx_0\in\R^p.  
\end{align*}
 The convergence of these systems is determined by the spectral radius of $\mC^{-1} \mA$. However, when these matrices depend on $\vx$, the spectral radius of each $\mC^{-1}_{\vx}\mA_{\vx}$ becomes insufficient to ensure the existence of solutions or the stability of dynamical systems described by~(\ref{eq:tilde-h}), specifically 
 \begin{align*}
   \vx_{k+1} = \mC^{-1}_{\vx_k}(\mA_{\vx_k} + \mC_{\vx_k})\vx_k + \mC^{-1}_{\vx_k}\vb_{\vx_k},\quad \vx_0\in\R^p.  
 \end{align*}
 Consequently, an alternative condition must be considered to address these challenges. In particular, we consider the following condition for some real number $c^*$:
\begin{align}
   \left\| \mC^{-1}_{\vx}(\mA_{\vx}+\mC_{\vx}) \right\|_{\infty}\leq c^*<1,\quad \forall \vx \in \gD \label{cond:avi-2}
\end{align}
for some set $\gD\subset\R^p$, which will be clarified further.
 Now, we have the result for existence of a solution of~(\ref{eq:pwa}):

\begin{lemma}\label{lem:avi-solution-existence}
    Suppose~(\ref{cond:avi-2}) holds with $\gD=\R^p$, and $\sup_{\vx\in\R^p}\left\|\mC^{-1}_{\vx}\vb_{\vx}\right\|_{\infty}<\infty$. If $\tilde{h}$ in~(\ref{eq:tilde-h}) is a continuous function, then there exists a solution of~(\ref{eq:pwa}).
\end{lemma}
\begin{proof}
For simplicity of the notation, let us denote $c=1-\sup_{\vx\in\R^p}\left\| \mC^{-1}_{\vx}(\mA_{\vx}+\mC_{\vx})\right\|_{\infty}$, and consider $H(\vx)=\vx+\alpha \tilde{h}(\vx)$ where $0<\alpha<\frac{1}{1-\sup_{\vx\in\R^p}\left\| \mC^{-1}_{\vx}(\mA_{\vx}+\mC_{\vx}) \right\|_{\infty}}$. Now, we have the following bound
\begin{align}
    &\left\|  (1-\alpha)\mI+\alpha \mC^{-1}_{\vx}(\mA_{\vx}+\mC_{\vx})   \right\|_{\infty}  \nonumber\\
    =& \max_{i\in[p]}\left| |(1-\alpha)+\alpha[\mC^{-1}_{\vx}(\mA_{\vx}+\mC_{\vx})]_{i,i}|+ \alpha \sum_{j\neq i}[[\mC^{-1}_{\vx}(\mA_{\vx}+\mC_{\vx})]_{ii}]_{i,j} \right| \nonumber\\
    \leq & 1-\alpha + \alpha \left\| \mC^{-1}_{\vx}(\mA_{\vx}+\mC_{\vx}) \right\|_{\infty} \nonumber\\
    <&1-\alpha c \label{ineq:1-alpha*c}
\end{align}
where the last two inequalities follows from the choice of $\alpha$.
    
     Then, we have
    \begin{align*}
        \left\| H(\vx) \right\|_{\infty}=& \left\|  \left((1-\alpha)\mI+\alpha\mC^{-1}_{\vx}(\mA_{\vx}+\mC_{\vx})  \right)\vx +\alpha\mC^{-1}_{\vx}\vb_{\vx}\right\|_{\infty}\\
        \leq & \left\|  (1-\alpha)\mI+\alpha \mC^{-1}_{\vx}(\mA_{\vx}+\mC_{\vx}) \right\|_{\infty}  \left\|\vx \right\|_{\infty}+\alpha \left\| \mC^{-1}_{\vx}\vb_{\vx} \right\|_{\infty}\\
        \leq & (1-\alpha c) \left\|\vx\right\|_{\infty}+\alpha \left\| \mC^{-1}_{\vx}\vb_{\vx} \right\|_{\infty}.
    \end{align*}

The last inequality follows from~(\ref{ineq:1-alpha*c}).

For $\vx$ such that $\left\|\vx\right\|_{\infty}\leq \frac{\sup_{\vx\in\R^p}\left\| \mC_{\vx}^{-1}\vb_{\vx} \right\|_{\infty}}{c}$, we have $\left\| H(\vx) \right\|_{\infty}\leq \frac{\sup_{\vx\in\R^p}\left\| \mC_{\vx}^{-1}\vb_{\vx} \right\|_{\infty}}{c} $. Therefore, $H$ is a self-map, and a continuous function from the assumption. Now, applying Brouwer's fixed point theorem in Lemma~\ref{lem:brouwer} in the Appendix, there exists a solution of~(\ref{eq:pwa}).    
\end{proof}

The uniqueness of the solution can be guaranteed with additional assumption of local lipshictzness of $\tilde{h}$:

\begin{lemma}\label{lem:avi-solution-unique-Lipschitz}
     Suppose $\mC_{\vx}$ and $\vb_{\vx}$ are bounded constant matrix and vector, respectively. If $\tilde{h}$ is a locally Lipschitz function and  $ \sup_{\vx\in\gD_{\tilde{h}}}\left\|\mC_{\vx}^{-1}(\mA_{\vx}+\mC_{\vx})\right\|_{\infty}<1$ where $\gD_{\tilde{h}}$ is the set of differentiable points of $\tilde{h}$, then a solution of~(\ref{eq:pwa}) exists and is unique.
\end{lemma}

\begin{proof}
    For simplicity, let us denote $c:=\sup_{\vx\in\gD_{\tilde{h}}}\left\|\mC_{\vx}^{-1}(\mA_{\vx}+\mC_{\vx})\right\|_{\infty}<1$.

    By a version of Lebourg's mean value theorem in Lemma~\ref{lem:lebourg-mvt-multi-dimension} in Appendix Section~\ref{app:sec:fine-property-function}, we have
        \begin{align*}
    &\left\| \mC^{-1}_{\vx}(\mA_{\vx}+\mC_{\vx})^{-1}\vx-\mC^{-1}_{\vy}(\mA_{\vy}+\mC_{\vy})^{-1}\vy  \right\|_{\infty}\leq c\left\|\vx-\vy \right\|_{\infty}.
    \end{align*}
    
    The Banach fixed point theorem in Lemma~\ref{lem:banach} in the Appendix ensures the uniqueness and existence of the solution.
\end{proof}

\begin{remark}
    Suppose there exists a norm $\left\|\cdot\right\|$ such that $\sup_{\vx\in\R^p}\left\|\mC_{\vx}^{-1}(\mA_{\vx}+\mC_{\vx})\right\|<1$, which can be guaranteed if the joint spectral radius is smaller than one~\citep{rota1960note} and $\{\mC_{\vx}^{-1}(\mA_{\vx}+\mC_{\vx}) : \vx \in\R^p\}$ is a finite set. Then, we can replace the infinity norm in~(\ref{cond:avi-2}) with this common norm. We refer to Definition~\ref{def:jsr} and Lemma~\ref{lem:jsr} in the Appendix~\ref{app:prelim} for further details.
\end{remark}

\subsection{Discussion on the condition~(\ref{cond:snrdd-2}) and~(\ref{cond:avi-2})}

From the above discussion, we can see that the two important assumptions are conditions (\ref{cond:snrdd-2}) and (\ref{cond:avi-2}). Let us discuss the scenario when both conditions are satisfied.

\begin{proposition}\label{prop:snrdd-splitting-contraction}
    Suppose for all $\vx\in\gD$, $\mC_{\vx}$ is a diagonal matrix such that $0<\sigma_{\min}<\lambda_{\min}(\mC_{\vx})$ and $\lambda_{\max}(\mC_{\vx})<\sigma_{\max}<\infty$ for some positive constants $\sigma_{\min}$ and $\sigma_{\max}$.
    
    \begin{enumerate}
        \item     Suppose~(\ref{cond:snrdd-2}) holds, and $\mA_{\vx}+\mC_{\vx}$ has non-negative diagonal elements for all $\vx\in\gD$, then,~(\ref{cond:avi-2}) holds.
        \item     If~(\ref{cond:avi-2}) holds, then~(\ref{cond:snrdd-2}) holds.
    \end{enumerate}
\end{proposition}

\begin{proof}
    From~(\ref{cond:snrdd-2}), for some $\kappa>0$, we have the following: 
    \begin{align*}
     -\kappa >&[\mA_{\vx}]_{i,i}+\sum_{j\in[p]\setminus\{i\}}|[\mA_{\vx}]_{i,j}|\\
     =&-[\mC_{\vx}]_{i,i}+[\mA_{\vx}+\mC_{\vx}]_{i,i}+\sum_{j\in[p]\setminus\{i\}}|[\mA_{\vx}+\mC_{\vx}]_{i,j}|\\
     =&[\mC_{\vx}]_{i,i}\left(-1+[\mC_{\vx}]^{-1}_{i,i}\sum_{j=1}^p \left| [\mA_{\vx}+\mC_{\vx}]_{i,j} \right|  \right)   .
    \end{align*}
    The first equality follows since $\mC_{\vx}$ is a diagonal matrix. The last equality follows from the fact that  the diagonal elements for $\mA_{\vx}+\mC_{\vx}$ are non-negative. As $[\mC_{\vx}]_{i,i}>0$, we have the following result:
    \begin{align*}
            0<\frac{\kappa}{\sigma_{\max}}\leq \sup_{\vx\in\gD}\left(1-\max_{i\in[p]}[\mC_{\vx}]^{-1}_{i,i}\sum_{j=1}^p \left| [\mA_{\vx}+\mC_{\vx}]_{i,j} \right|  \right)=1-\sup_{\vx\in\gD}\left\| \mC_{\vx}^{-1}(\mA_{\vx}+\mC_{\vx}) \right\|_{\infty}.
    \end{align*}

    This proves the first statement.

    Now, let us prove the second statement. Note that from the condition~(\ref{cond:avi-2}), for some $\omega>0$,
    \begin{align*}
          -\omega >&\max_{i\in[p]}\frac{\sum_{j=1}^p\left| [\mA_{\vx}+\mC_{\vx}]_{i,j} \right|}{[\mC_{\vx}]_{i,i}}-1\\
        = & \max_{i\in[p]} \frac{1}{[\mC_{\vx}]_{i,i}}\left(\sum_{j=1}^p\left| [\mA_{\vx}+\mC_{\vx}]_{i,j} \right|-[\mC_{\vx}]_{i,i} \right)\\
        \geq & \max_{i\in[p]}\frac{1}{\sigma_{\min}} \left(\sum_{j=1}^p\left| [\mA_{\vx}+\mC_{\vx}]_{i,j} \right|-[\mC_{\vx}]_{i,i} \right).
    \end{align*}

    The first inequality follows from the assumption in~(\ref{cond:avi-2}), and the  last inequality follows from the condition $\max_{i\in[p]
    }[\mC_{\vx}]_{i,i}>\sigma_{\min}$. 

    Now, we can check that 
    \begin{align*}
    -\sigma_{\min}\omega > \sum_{j=1}^p\left| [\mA_{\vx}+\mC_{\vx}]_{i,j} \right|-[\mC_{\vx}]_{i,i} \geq \sum^p_{j=1} \sum^p_{j\in[p]\setminus\{i\}}|[\mA_{\vx}]_{i,j}|+\mA_{\vx}.
    \end{align*}
    where the last inequality follows from the fact that $\mC_{\vx}$ is a diagonal matrix.

    Therefore, from the definition of $S_i(\cdot)$ in Definition~\ref{def:snrdd}, taking supremum and maximum over the above inequality, we get

    \begin{align*}
       \sup_{\vx\in\gD}\max_{i\in[p]}S_i(\mA_{\vx}) \leq \sup_{\vx\in\gD}\max_{i\in[p]}\left(\sum_{j=1}^p\left| [\mA_{\vx}+\mC_{\vx}]_{i,j} \right|-[\mC_{\vx}]_{i,i} \right) \leq -\sigma_{\min}w<0,
    \end{align*}
which proves the second statement.

\end{proof}


\section{Auxiliary Details}\label{app:prelim}

\subsection{Fine properties of function}\label{app:sec:fine-property-function}

\begin{definition}[Locally Lipschitz function~\cite{clarke1981generalized}]\label{def:locally-Lipschitz}
A function $f:\R^n\to\R^n$ is said to be locally Lipschitz, if for $\vx\in\R^n$, there exists a constant $L$ an $\delta$ such that
\begin{align*}
    \left\|\vx-\vx_0\right\|<\delta\Rightarrow \left\|f(\vx)-f(\vx_0)\right\|\leq L \left\|\vx-\vx_0\right\|
\end{align*}
\end{definition}

\begin{lemma}[Rademacher's theorem, page 810 in~\cite{evans2018measure}]\label{lem:rademacher}  
Let $f:\R^n\to\R^n$. If $f$ is a locally lispcthiz function, then $f$ is differentiable almost everywhere.
\end{lemma}

\begin{definition}[Generalized directional derivative]
The generalized directional derivative of the locally Lipschitz function $f:\R^n\to\R$ at the point $\vu\in\R^n$ in the direction $\vv\in\R^n$ is defined by
\begin{align*}
    f^o(\vu;\vv)= \limsup_{\vw\to\vu,t\to 0+} \frac{f(\vw+t\vv)-f(\vw)}{t}
\end{align*}
\end{definition}

\begin{definition}[Clarke subdifferential, page 54 in~\cite{clarke1981generalized}]
Let $f:\R^n\to\R$ be a locally Lipschitz function. The Clarke subdifferential $\partial_C f(u)$ of $f$ at a point $u\in\R^n$ is defind as the following:
\begin{align*}
\partial_C f(\vu) = \left\{ \vv \in \R^n : \vv^{\top}\vy \leq f^o(\vu;\vv) ,\;\forall \vy\in \R^n \right\}
\end{align*}
When $f$ is locally Lipschitz, then
\begin{align*}
    \partial_Cf(\vu)= \mathrm{conv} \left\{ \lim_{i\to\infty} \nabla f(\vx_i) : \text{$\vx_i\in\gD_f$ such that $\vx_i\to \vv$ and $\lim_{i\to\infty} \nabla f(\vx_i)$ exists }\right\}
\end{align*}
where $\mathrm{conv}(A)$ denotes convex hull of a set $A$, and $\gD_f$ is the differentiable points of $f$ and $\vx_i$ is a converging sequence to $\vv$.
\end{definition}

\begin{lemma}[Lebourg's mean value theorem, Theorem 2.4 in~\cite{clarke2008nonsmooth}]\label{lem:lebourg-mvt}
Consider a locally Lipschitz function $f:\R^n\to\R$. For $\vx,\vy\in \R^n$, there exists $\vv\in \{ t\vx+(1-t)\vy : t\in [0,1]\}$ such that
\begin{align*}
    f(\vx)-f(\vx) =& \vz^{\top}(\vx-\vy)
\end{align*}
where
\begin{align*}
\vz\in \partial f_C(\vv)  =& \mathrm{conv} \left\{ \lim_{i\to\infty} \nabla f(\vx_i) : \text{$\vx_i\in\gD_f$ such that $\vx_i\to \vv$ and $\lim_{i\to\infty} \nabla f(\vx_i)$ exists } \right\}.
\end{align*}
$\gD_f$ is the differentiable points of $f$ and $\{\vx_i\in\gD_f\}_{i=1}^{\infty}$ is a converging sequence to $\vv$.
\end{lemma}

\begin{lemma}\label{lem:lebourg-mvt-multi-dimension}
    Suppose $f:\R^n\to\R^n$ is a locally lipscthiz function and $ \left\| \nabla f(\vx) \right\|_{\infty} \leq f_{\max}$ for all the differentiable points $\vx$ for some positive real number $f_{\max}$. Then, the following holds:
    \begin{align*}
        \left\|f(\vx)-f(\vy) \right\|_{\infty}\leq f_{\max}\left\|\vx-\vy\right\|_{\infty}.
    \end{align*}
\end{lemma}
\begin{proof}
    Consider $\ve_i^{\top}f(\vx)$ for some $i\in[n]$. By Lebourg's mean value theorem in Lemma~\ref{lem:lebourg-mvt}, we have, for a basis vector in $\R^n$ whose $i$-th coordinate is one,
    \begin{align}
        \ve_i^{\top}(f(\vx)-f(\vy))=  \va_i^{\top} (\vx-\vy) 
    \end{align}
    for $\va_i\in\text{conv}\{ \lim_{k\to\infty}\nabla f(\vx_k)^{\top}\ve_i  :\vx_k\to \vv, \vx_k \in \gD_f\}$ where $\vv\in \{ t\vx+(1-t)\vy : t \in [0,1]\}$. We can find such $\va_i$ for all $i\in[n]$, and we have
    \begin{align*}
        f(\vx)-f(\vy) = \begin{bmatrix}
            \va_1^{\top}\\
                \vdots\\
            \va_n^{\top}
        \end{bmatrix}  (\vx-\vy).
    \end{align*}
    Taking the infinity norm on both sides, we get
    \begin{align}
        \left\| f(\vx)-f(\vy) \right\|_{\infty} \leq  &  \max_{i\in[n]}\left\|  \va_i^{\top}\right\|_{\infty} \left\|\vx- \vy \right\|_{\infty} \nonumber \\
        \leq  &  \left\| \sum_{j=1}^q \lambda_j \hat{f}_j \right\|_{\infty} \left\| \vx - \vy  \right\|_{\infty} \label{ineq:gmvt-1}
    \end{align} 
    where $\sum_{j=1}^q\lambda_j=1$. For $j\in[q]$, $\lambda_j\geq 0$, and $\hat{f}_j=\lim_{k\to\infty} \ve_i^{\top}\nabla f(\vx^j_k)$ for some sequence $\{\vx^j_k\in\gD_f\}_{k=1}^{\infty}$. Note that we have
    \begin{align*}
        \left\|\sum_{j=1}^q  \lambda_j \hat{f}_j \right\|_{\infty}   =&  \lim_{k\to\infty}\left\| \sum_{j=1}^q \ve_i^{\top}\nabla f(\vx_k^j)\right\|_{\infty}\\
        \leq &  \lim_{k\to\infty} \sum_{j=1}^n \lambda_j \left\|\nabla f(\vx^j_k)\right\|_{\infty}\\
           \leq & \sum^n_{j=1}\lambda_j f_{\max} \\
           = & f_{\max} .
    \end{align*}
    Applying this result to~(\ref{ineq:gmvt-1}) yields the desired result. 
\end{proof}

\begin{lemma}\label{lem:local-Lipschitz}
    A function $f:\R^n\to\R^n$ is locally Lipschitz if and only if $f$ is Lipschitz on every compact subset $K\subset\R^n$.
\end{lemma}
\begin{proof}
The necessity part is an immediate consequence of the definition of local Lipschitz continuity. We now establish the converse implication. Suppose the local Lipschitzness holds with some norm $\left\|\cdot\right\|$ and $f$ is not Lipschitz on some compact set $K$. Then, there exists some $\vx,\vy\in K$ such that $\frac{\left\| f(\vx)-f(\vy) \right\|}{||\vx-\vy||}>C$ for any $C\geq 0$. Therefore, there exist a sequence $\{ (\vx_n,\vy_n) \in K\times K \}_{n=1}^{\infty}$ such that $\frac{\left\| f(\vx_n)-f(\vy_n) \right\|}{||\vx_n-\vy_n||}\to \infty$. From the compactness of $K$, there exist a convergent subsequence $\{(\vx_{k_n},\vy_{k_n})\}_{n=1}^{\infty}$ converging to $(\tilde{\vx},\tilde{\vy})$. Moreover, as continuous function is bounded on compact set, we should have $||\vx_{k_n}-\vy_{k_n}||\to0$. This contradicts the fact that $f$ is locally Lipschitz at $\tilde{\vx}$, and this proves the reverse direction.
\end{proof}

\begin{lemma}[Brouwer's fixed point theorem~\citep{brouwer1911}]\label{lem:brouwer}
    Let $\gB_R:=\{\vx\in\R^n : \left\|\vx\right\|<R  \}$ be an open ball in $\R^n$ centered at the origin and of radius $R$ with some norm $\left\|\cdot\right\|$. If $f:\gB_R\to\gB_R$ is a continuous function, then, $f$ has a fixed point, i.e., a solution of $f(\vx)=\vx$. 
\end{lemma}

\begin{lemma}[Banach fixed point theorem~\citep{banach1922operations}]\label{lem:banach}
 Consider a mapping $f:\R^n\to\R^n$. Suppose there exists a norm $\left\|\cdot \right\|$ such that $\left\| f(\vx)-f(\vy)\right\|< C \left\|\vx-\vy\right\|$ where $C\in(0,1)$. Then, there exists a unique point $\vx^*\in\R^n$ such that $f(\vx^*)=\vx^*$.
\end{lemma}

\subsection{Matrix properties}




\begin{lemma}[Lemma 7 in~\cite{davydov2024non}]\label{lem:one-sided-snrdd-equivalence}
    Suppose the map $f$ is locally Lipschitz. Then, the following two conditions are equivalent
    \begin{align*}
\text{$\max_{i\in[p]}S_i(\nabla f(\vx)) \leq -c$ for $\vx\in \gD_f$}        \iff \text{ $f$ is one-sided Lipschitz with constant $-c$ in $\gD_f$}
    \end{align*}
    where $\gD_f$ is the set of differentiable points of $f$ and $S_i(\cdot)$ is defind in Definition~\ref{def:snrdd}.
\end{lemma}

Note that only Lipschitz continuity is required in the above lemma instead of differentiablity of $f$ in $\R^n$. By Rademacher theorem, a Lipschitz continuous function is differentiable almost everywhere.



Now, let us briefly explain the concept of joint spectral radius~\citep{rota1960note}, which is defined as follows:

\begin{definition}[Joint spectral radius~\citep{rota1960note}]\label{def:jsr}
\end{definition}
Given a set of matrix $\{\mA_i\in\R^{n\times n}\}_{i=1}^m$, the joint spectral radius is defined as
\begin{align*}
    \rho(\mA_1,\cdots,\mA_m)= \lim_{k\to\infty}\max_{\sigma\in \{1,2,\dots,m\}^k}\left\| \mA_{\sigma_k}\cdots\mA_{\sigma_2}\mA_{\sigma_1} \right\|^{1/k}.
\end{align*}

\begin{lemma}[\cite{rota1960note}]\label{lem:jsr}
    Given a set of matrix $\{\mA_i\in\R^{n\times n}\}_{i=1}^m$, if the joint spectral radius is smaller than one, then there exists a norm $\left\|\cdot \right\|$ such that $\left\|\mA_i\right\|<1$ for all $i\in[m]$.
  \end{lemma}




\begin{lemma}[Gerschgorin circle theorem~\citep{horn2012matrix}]\label{gersgorin-circel-theroem}
Let \( \mA \in \mathbb{R}^{n\times n}\) and \( R_i (\mA) = \sum\limits_{j\in [n]\setminus \{i\} } [\mA]_{i,j} \). Consider the Gerschgorin circles
\begin{align*}
    \{ z\in \mathbb{C} | : | z- [\mA]_{i,i} | \leq R_i (\mA) \} ,\quad i=1,\dots,n.
\end{align*}
The eigenvalues of \(\mA\) are in the union of Gerschgorin discs
\begin{align*}
    G(\mA) = \cup^n_{i=1}  \{ z\in \mathbb{C} | : | z- [\mA]_{i,i} | \leq R_i (\mA) \} .
\end{align*}
\end{lemma}

\begin{definition}[Hurwtiz matrix~\citep{khalil2002nonlinear}]
A matrix $\mA\in\R^{n\times n}$ is said to be a Hurwitz matrix if all of its eigenvalues has negative real part.
\end{definition}

\begin{lemma}\label{lem:snrdd-then-hurwitz}
    An SNRDD matrix is a Hurwitz matrix.
\end{lemma}

\begin{proof}
    The proof directly follows from Gerschgorin circle theorem in Lemma~\ref{gersgorin-circel-theroem}.
\end{proof}


\subsection{Types of policies and Markov chain}\label{sec:choice-of-policy}

\textbf{$\eps$-greedy policy:} Let $\gA^* =arg\max_{a\in\gA} \vphi(s,a)^{\top}\vtheta$.
\begin{align*}
    \pi_{\vtheta}^{\eps}(a\mid s)=&\begin{cases}
         \frac{1}{|\gA^*|}-\frac{\eps}{|\gA^*|}  &\text{if} \quad a\in \gA^* \\
                    \frac{\eps}{|\gA|-|\gA^*|}      &\text{if} \quad a \notin \gA^*
    \end{cases} 
\end{align*}

\textbf{$\eps$-softmax policy} : Given a positive real number, $\tau$, which is so-called a temperature parameter, the $\eps$-softmax policy policy is defined as 
\begin{align*}
    \pi_{\vtheta}(a\mid s) = \frac{\exp(\tau \vphi(s,a)^{\top}\vtheta)}{\sum_{u\in\gA}\exp( \tau \vphi(s,u)^{\top}\vtheta)} ,\quad \forall (s,a)\in\gS\times \gA
\end{align*}

\textbf{Tamed Gibbs Policy~\citep{meyn2024projected}}
A $(\eps,\kappa_0)$-tamed Gibbs policy defined as $\pi_{\vtheta}(a\mid s) = \frac{\exp(-\tau_{\vtheta}\vphi(s,a)^{\top}\vtheta)}{\sum_{u\in\gA}\exp\left(-\tau_{\vtheta}\vphi(s,u)^{\top}\vtheta \right)}$ where
\begin{align*}
    \tau_{\vtheta} ( a\mid s) = \begin{cases}
          \frac{\kappa_0}{\left\|\vtheta\right\|_2}  & \left\|\vtheta\right\|_2 \geq 1\\
          \frac{\kappa_0}{2} & \text{else}
    \end{cases}.
\end{align*}

The following lemma is from~\cite{perkins2002convergent}:

\begin{lemma}\label{lem:stationary-distribution-local-Lipschitz}
If the behavior policy $\beta_{\vtheta}$ satisfying Assumption~\ref{assumption:statinoary-distribution} is locally Lipschitz, then, its corresponding stationary distribution $\mu_{\beta_{\vtheta}}$ is also locally Lipschitz. 
\end{lemma}
\begin{proof}

    For simplicity of the proof, let $\mP_{\beta_{\vtheta}}\in\R^{|\gS||\gA|\times|\gS||\gA|}$ and $\vmu_{\beta_{\vtheta}}\in\R^{|\gS||\gA|}$ such that
\begin{align*}
    [\mP_{\beta_{\vtheta}}]_{(s-1)|\gA|+a,(x-1)|\gA|+u} = \gP(x\mid s,a)\beta_{\vtheta}(u\mid x),\quad [\vmu_{\beta_{\vtheta}}]_{(s-1)|\gA|+a}= \mu_{\beta_{\vtheta}}(a\mid s).
\end{align*}
From local lipschitszness of $\beta_{\vtheta}$, there exists $\delta$ and $L$ such that for $\vtheta^{\prime}$ satisfying $|\beta_{\vtheta}(a\mid s)-\beta_{\vtheta^{\prime}}(a\mid  s)|$ then, $||\leq L\left\| \vtheta-\vtheta^{\prime} \right\|$ for some norm $\left\|\cdot\right\|$.

Now, note that the following holds~\citep{seneta1993sensitivity}: 
    \begin{align*}
        \vmu_{\beta_{\vtheta^{\prime}}}^{\top}-\vmu_{\beta_{\vtheta}}^{\top} = \vmu_{\beta_{\vtheta^{\prime}}}^{\top}(\mP_{\beta_{\vtheta^{\prime}}}-\mP_{\beta_{\vtheta}})  (\mI-\mP_{\beta_{\vtheta}}+\bm{1}\vmu_{\beta_{\vtheta}}^{\top})^{-1}.
    \end{align*}
    Therefore, taking norm on each sides,
    \begin{align}
        \left\|  \vmu_{\beta_{\vtheta^{\prime}}}^{\top}-\vmu_{\beta_{\vtheta}}^{\top}\right\|_1 \leq & \left\| (\mI-\mP_{\beta_{\vtheta}}+\bm{1}\vmu_{\beta_{\vtheta}}^{\top})^{-1}\right\|_1  \left\| \mP_{\beta_{\vtheta}}-\mP_{\beta_{\vtheta^{\prime}}}\right\|_1 \nonumber \\
        = & \left\|\sum^{\infty}_{k=0}(\mP_{\beta_{\vtheta}}-\bm{1}\vmu^{\top}_{\beta_{\vtheta}})^k \right\|_1\left\| \mP_{\beta_{\vtheta}}-\mP_{\beta_{\vtheta^{\prime}}}\right\|_1 \nonumber\\
        \leq & C_{\mu_{\beta_{\vtheta^{\prime}}}}\left\|\sum^{\infty}_{k=0}(\mP_{\beta_{\vtheta}}-\bm{1}\vmu^{\top}_{\beta_{\vtheta}})^k \right\|_u\left\| \mP_{\beta_{\vtheta}}-\mP_{\beta_{\vtheta^{\prime}}}\right\|_1  \nonumber\\
        \leq & \frac{C_{\mu_{\beta_{\vtheta}}}}{1-\left\|  \mP_{\beta_{\vtheta}}-\bm{1}\vmu^{\top}_{\beta_{\vtheta}}\right\|_u}  \left\| \mP_{\beta_{\vtheta}}-\mP_{\beta_{\vtheta^{\prime}}}\right\|_1 \nonumber\\
        \leq & \frac{C_{\mu_{\beta_{\vtheta}}}}{1-\left\|  \mP_{\beta_{\vtheta}}-\bm{1}\vmu^{\top}_{\beta_{\vtheta}}\right\|_u} L\left\|\vtheta-\vtheta^{\prime} \right\| \nonumber
    \end{align}
    where $\left\|\cdot\right\|_{u}$ is a norm such that $\left\| \mP_{\beta_{\vtheta}}-\bm{1}\vmu^{\top}_{\beta_{\vtheta}}\right\|_{u}<1$ which exists as $\rho(\mP_{\beta_{\vtheta^{\prime}}}-\bm{1}\vmu^{\top}_{\beta_{\vtheta^{\prime}}})<1$.  $C_{\mu_{\beta_{\vtheta}}}$ is a scalar such that $\left\|\cdot\right\|\leq C_{\mu_{\beta_{\vtheta}}}\left\| \cdot\right\|_u$ which exists by equivalence of norm. The second inequality follows from sum of geometric series. Therefore, local liopschitzness of $\mu_{\beta_{\vtheta}}(a\mid s)$ follows.
\end{proof}

\section{Stochastic approximation}\label{sec:sa}

Let us consider a stochastic approximation scheme~\citep{robbins1951stochastic}:
\begin{align*}
    \vx_{k+1} = \vx_k + \alpha_k \left(f(\vx_k) + \veps_k \right)
\end{align*}
where $f:\R^n\to\R^n$ is continuous function, $\veps_k$ is a Martingale difference sequence and $\alpha_k\in [0,1]$ is the step-size. 

\begin{definition}[Martingale difference sequence]\label{def:mds}
    Consider a sequence of random variables $\veps_0,\veps_1,\dots$ and let $\sigma$-fields $\gF_k=\sigma(\vx_0,\veps_1,\dots,\veps_k)$. If $\E\left[ \veps_{k+1} \middle| \gF_k \right]=0$ and $\E[ \left\|\veps_{k+1} \right\|^2  | \gF_k ]<\infty$ almost surely for the $\sigma$-fields $\gF_k$, $\veps_k$ is called a Martingale difference sequence.
\end{definition}

The ODE counterpart can characterize the stability of stochastic approximation scheme:
\begin{align*}
    \dot{\vx}_t = f(\vx_t), \quad \vx_0\in\R^n, t\geq 0.
\end{align*}
\begin{assumption}\label{borkar_meyn_assumption} 
1. The mapping \(f: \R^n \rightarrow \R^n\) is globally Lipschitz continuous, and there exists a function \(f_{\infty} : \R^n \rightarrow \R^n\) such that
    \begin{equation}
        \lim_{c\rightarrow\infty} \frac{f(c\vx)}{c} = f_{\infty}(\vx) , \quad \forall{\vx} \in \R^n.    
    \end{equation}
2. The origin in \(\R^n\) is a globally asymptotically stable equilibrium for the ODE \(\dot{\vx}_t = f_{\infty}(\vx_t)\).
\newline
\newline
3. There exists a globally asymptotically stable equilibrium \(\vx^*\in\mathbb{R}^n\) for the ODE \( \dot{\vx}_t=f(\vx_t)\) , i.e., \(\vx_t \rightarrow \vx^*\) as \( t\rightarrow\infty\).
\newline
\newline
4. The sequence \(\{ \veps_k,\gG_k  \}_{k\geq 1}  \) where \( \gG_k \) is sigma-algebra generated by \(\{(\vx_i,\veps_i, k\geq i \}\), is a Martingale difference sequence. In addition , there exists a constant \( 
C_0 < \infty \) such that for any initial \(  \theta_0 \in \R^n \) , we have \(\E [|| \veps_{k+1} ||^2 | \gG_k ] \leq C_0 (1+|| \vx_k ||^2), \forall{k}\geq 0  \).
\newline
\newline
5. The step-sizes satisfies the Robbins-Monro condition~\citep{robbins1951stochastic} :
\begin{align*}
    \sum\limits^{\infty}_{k=0}\alpha_k = \infty,\quad \sum\limits^{\infty}_{k=0}\alpha_k^2 < \infty.
\end{align*} 
\end{assumption}

\begin{lemma}[Borkar and Meyn Theorem,~\cite{borkar2000ode}]\label{lem:borkar-meyn}
    Suppose there exists unique $\vx^*\in\R^n$ such that $f(\vx^*)=\bm{0}$ and Assumption~\ref{borkar_meyn_assumption} holds. Then, $\vx_k\to\vx^*$ with probability one.
\end{lemma}

\section{Omitted proofs in main manuscript}\label{sec:omitted-proofs}

\begin{lemma}\label{lem:eta>3->SNRDD}
    If $\eta>3$, and $\left\|\vphi(s,a)\right\|_{\infty}\leq \frac{1}{\sqrt{p}}$ for all $(s,a)\in\gS\times\gA$, then $\gamma\mPhi^{\top}\mD_d\mP\mPi_{\pi^g_{\vtheta}}\mPhi-(\eta\mI+\mPhi^{\top}\mD_d\mPhi)$ is SNRDD for all $\vtheta\in\R^p$.
\end{lemma}

\begin{proof}
        For $i\in\{1,2,\dots,p\}$, let us consider $S_i(\gamma\mPhi^{\top}\mD_d\mP\mPi_{\pi^g_{\vtheta}}\mPhi-(\eta\mI+\mPhi^{\top}\mD_d\mPhi))$ which is defined in Definition~\ref{def:snrdd}:
    \begin{align*}
    & \left( -\eta   -[\mPhi^{\top}\mD_d\mPhi]_{i,i}^2 + \gamma[\mPhi^{\top}\mD_d\mP\mPi_{\pi^g_{\vtheta}} \mPhi ]_{i,i} +\sum_{j\in\{1,2,\dots,p\}\setminus \{i\}} \left| [-\mPhi^{\top}\mD_d\mPhi+\gamma\mPhi^{\top}\mD_d\mP\mPi_{\pi^g_{\vtheta}}\mPhi]_{i,l}\right| \right)\\
       =& -\eta - \sum_{(s,a)\in\gS\times\gA}d(s,a)\left([\vphi(s,a)]_i^2-\gamma \sum_{s^{\prime}\in\gS}\gP(s^{\prime}\mid s,a)[\vphi(s,a)]_i \left[\sum_{u\in\gA} \pi_{\vtheta}(u \mid s^{\prime}) \vphi(s^{\prime},u)\right]_i\right) \\
       &+  \sum_{j\in\{1,2,\dots,p\}\setminus \{i\} }\sum_{(s,a)\in\gS\times\gA} \left|  d(s,a) \sum_{s^{\prime}\in\gS}\gP(s^{\prime}\mid s,a) [\vphi(s,a)]_i \left([\vphi(s,a)]_j-\gamma \left[\sum_{u\in\gA} \pi_{\vtheta}(u \mid s^{\prime}) \vphi(s^{\prime},u)\right]_j  \right) \right|\\
       \leq & -\eta + \sum_{(s,a)\in\gS\times\gA}d(s,a)\left(\left\|\vphi(s,a)\right\|_{\infty}^2 + \gamma \sum_{s^{\prime}\in\gS}\gP(s^{\prime}\mid s,a)\left\|\vphi(s,a)\right\|_{\infty}\left\| \sum_{u\in\gA} \pi_{\vtheta}(u\mid s^{\prime}) \vphi(s^{\prime},u) \right\|_{\infty}\right) \\
        &+ \sum_{j\in\{1,2,\dots,p\}\setminus \{i\} }\sum_{(s,a)\in\gS\times\gA} d(s,a) \sum_{s^{\prime}\in\gS}\gP(s^{\prime}\mid s,a) \left\|\vphi (s,a)\right\|_{\infty}\left(\left\|\vphi (s,a)\right\|_{\infty}+ \gamma \sum_{u\in\gA}\pi_{\vtheta}(u\mid s^{\prime})\left\|\vphi(s^{\prime},u)\right\|_{\infty} \right)\\
        \leq & -\eta + \sum_{(s,a)\in\gS\times\gA} d(s,a)\left( \frac{1}{p}+\gamma \sum_{s^{\prime}\in\gS}\gP(s^{\prime}\mid s,a) \frac{1}{p}\right)\\
        &+ \sum_{j\in \{1,2,\dots,p\}\setminus \{i\}}\sum_{(s,a)\in\gS\times\gA}d(s,a)\sum_{s^{\prime }\in\gS} \gP(s^{\prime}\mid s,a) \left( \frac{2}{p} \right)\\
        \leq & -\eta + \frac{2}{p}+2\\
        \leq & - \eta +3.
\end{align*}

The first inequality follows from the fact that $|[\vphi(s,a)]_i|\leq \left\| \vphi(s,a) \right\|_{\infty}$ for all $(s,a)\in\gS\times\gA$ and $i\in\{1,2,\dots,p\}$ and using triangle inequality. The second inequality follows from the assumption that $\left\|\vphi(s,a)\right\|_{\infty}\leq\frac{1}{\sqrt{p}}$ for all $(s,a)\in\gS\times\gA$. Therefore, $\eta>3$ is sufficient for our goal.

\end{proof}

\subsection{Proof of Theorem~\ref{prop:pbe-existence:1}}\label{app:prop:pbe-existence:1}

Now, let us present the proof of Theorem~\ref{prop:pbe-existence:1}:
\begin{proof}
    The first statement follows from Lemma~\ref{lem:continuity-fixed-point} in the Appendix, which generalizes Theorem~\ref{prop:pbe-existence:1}. The proof outline is as follows : We can check that \(\vx + \alpha F(\vx)\) is a self-map, meaning it maps \(\gC\) onto itself, where \(\gC\) is a compact set, and the function $F$ is continuous. 
    
    The reward is bounded by the assumption, and using the condition in~(\ref{cond:snrdd}), we can show that \(\vx + \alpha F(\vx)\) is a self-map. Moreover, using a continuous behavior and target policy, the function $F$ is continuous. Then, we can apply the Brouwer's fixed point theory in Lemma~\ref{lem:brouwer} in the Appendix.
    
    The second statement follows from Lemma~\ref{lem:snrdd-uniqueness-existence}, which applies the result of~\cite{davydov2024non} that for locally Lipschitz function with SNRDD condition, the uniqueness of the solution is guaranteed. The only condition we need to check is the local Lipschitzness of $F$. The stationary distribution $\mu_{\beta_{\vtheta}}$ is locally Lipschitz from Lemma~\ref{lem:stationary-distribution-local-Lipschitz} in the Appendix. As product of locally lipscthiz functions are still locally lispchitz, which can be verified using Lemma~\ref{lem:local-Lipschitz} in the Appendix, $F$ is a locally function. Therefore, we can now apply Lemma~\ref{lem:snrdd-uniqueness-existence} in the Appendix.
\end{proof}

\subsection{Proof of Proposition~\ref{prop:pbe-existence:2}}\label{app:prop:pbe-existence:2}
\begin{proof}
The first statement is a specific case of Lemma~\ref{lem:avi-solution-existence} in the Appendix, a generalized version of the first statement. The idea of the proof of Lemma~\ref{lem:avi-solution-existence} is the following : Consider the scenario when~(\ref{suff-cond:avi-convrge:1}) holds. Multiplying $\mPhi$ on both sides of~(\ref{pbe:2}), we get
    \begin{align*}
        \mPhi\vtheta = \mPhi (\mPhi^{\top}\mD_{\mu_{\beta_{\vtheta}}}\mPhi)^{-1}\left(\gamma\mPhi^{\top}\mD_{\mu_{\beta_{\vtheta}}}\mP\mPi_{\pi_{\vtheta}}\mPhi\vtheta+\mPhi^{\top}\mD_{\mu_{\beta_{\vtheta}}}\mR \right) .
    \end{align*}
    Let $\beta_{\vtheta}=\beta_{\mPhi\vtheta}$ and $\pi_{\vtheta}=\pi_{\mPhi\vtheta}$. Denote $\vy =  \mPhi\vtheta  $ and $\tilde{F}(\vy):=\mPhi (\mPhi^{\top}\mD_{\mu_{\beta_{\vy}}}\mPhi)^{-1}\left(\gamma\mPhi^{\top}\mD_{\mu_{\beta_{\vy}}}\mP\mPi_{\pi_{\vy}}\vy+\mPhi^{\top}\mD_{\mu_{\beta_{\vy}}}\mR \right)$. Now, it is sufficient to investigate the solution of the equation $\vy=\tilde{F}(\vy)$. We can check that $\vy+\alpha \tilde{F}(\vy)$ is a self-map for some small enough $\alpha$, and a continuous map, and we can apply Brouwer's fixed point theorem in Lemma~\ref{lem:brouwer} in the Appendix. 
    Therefore, we can apply Lemma~\ref{lem:avi-solution-existence} in Appendix Section~\ref{sec:fixed-point}. The same argument holds when we consider the condition~(\ref{suff-cond:avi-convrge:2}).

    The second statement is a specific case of Lemma~\ref{lem:avi-solution-unique-Lipschitz} in the Appendix~\ref{sec:fixed-point}. The proof relies on a version of Lebourg's mean value theorem in Lemma~\ref{lem:lebourg-mvt-multi-dimension} in the Appendix.
\end{proof}

\subsection{Proof of Proposition~\ref{prop:pbe-splitting-contraction}}\label{app:prop:pbe-splitting-contraction}
\begin{proof}
A generalized version of proof is provided in Proposition~\ref{prop:snrdd-splitting-contraction} in Appendix Section~\ref{sec:fixed-point}. The proof can be established using the definition of the infinity norm and SNRDD as given in Definition~\ref{def:snrdd}. 
\end{proof}

\subsection{Proof of Lemma~\ref{lem:q-learning-osl}}\label{app:lem:q-learning-osl}
\begin{proof}
    Let us provide a proof of first statement, the case of asynchronous tabular Q-learning. For $i\in \gI_{\infty}(\mQ-\tilde{\mQ})$,
\begin{align*}
&[\mQ-\tilde{\mQ}]_{i}[F_{\text{AsyncQ}}(\mQ)-F_{\text{AsyncQ}}(\tilde{\mQ})]_{i}\\=& [\mQ-\tilde{\mQ}]_{i} [\gamma\mD_d\mP (\mPi_{\pi_{\mQ}}\mQ-\mPi_{\pi_{\tilde{\mQ}}}\tilde{\mQ}) - \mD_d(\mQ-\tilde{\mQ})]_{i}\\
=&[\mQ-\tilde{\mQ}]_{i} \left( \gamma [\mD_d]_{i,i} \sum_{j\in[|\gS|]}[\mP]_{i,j}\left(\max_{u\in\gA}[\mQ]_{(j-1)|\gA|+u}-\max_{u\in\gA}[\tilde{\mQ}]_{(j-1)|\gA|+u}\right)- [\mD_d]_{i,i}[\mQ-\tilde{\mQ}]_{i} \right)\\
\leq & -[\mD_d]_{i,i}\left|[\mQ-\tilde{\mQ}]_i\right|^2+|[\mQ-\tilde{\mQ}]_i|\left( \gamma [\mD_d]_{i,i} \sum_{j\in\gS} [\mP]_{i,j} \max_{u\in\gA} |[\mQ]_{(j-1)|\gA|+a}-[\tilde{\mQ}]_{(j-1)|\gA|+a} |\right)\\
\leq & (\gamma-1) [\mD_d]_{i,i} \left| [\mQ-\tilde{\mQ}]_i\right|^2\\
\leq & (\gamma-1) d_{\min} \left| [\mQ-\tilde{\mQ}]_i\right|^2.
\end{align*}    
The second last line follows from the non-expansiveness of the max-operator and $|[\mQ-\tilde{\mQ}]_{(s-1)|\gA|+a}|\leq |[\mQ-\tilde{\mQ}]_i |$ for all $(s,a)\in\gS\times\gA$ since $i\in \gI_{\infty}(\mQ-\tilde{\mQ})$. This proves the first statement.

Now, let us prove the second statement, the case for linear Q-learning. For $\vtheta,\tilde{\vtheta}\in\R^p$ and $i\in\gI_{\infty}(\vtheta-\tilde{\vtheta})$, from Lebourg's mean value theorem in Lemma~\ref{lem:lebourg-mvt} in the Appendix, we have
\begin{align*}
 [F_{\mathrm{linear}}(\vtheta)-F_{\mathrm{linear}}(\tilde{\vtheta})]_i = \ve_i^{\top} \mA(\vtheta-\tilde{\vtheta})
\end{align*}
where $\ve_i$ is the unit vector in $\R^p$. For $\vv\in \{ t\vx + (1-t)\vy : t\in[0,1] \}$ $\mA\in \text{conv}\{ \lim_{k\to\infty} \nabla F_{\mathrm{linear}}(\vx_k) : \vx_k \to \vv, \;  \vx_k\in \gD_{F_{\mathrm{linear}}} \}$ and $\gD_{F_{\mathrm{linear}}}$ is the differentiable points of $F_{\mathrm{linear}}$.  $\mA$ can be expressed as $\mA=\sum^q_{j=1} \lambda_j \lim_{k\to\infty}  \nabla F_{\mathrm{linear}}(\vx^j_k)$ for some $q\in\sN$, $\sum^q_{j=1}\lambda_j=1$ and for $j\in[q]$, $\lambda_j\geq 0$ and $\{\vx^j_k\}_{k=1}^{\infty}$ is a converging sequence to $\vv$. We have,
\begin{align*}
    &[\vtheta-\tilde{\vtheta}]_i[\mA(\vtheta-\tilde{\vtheta})]_i \\
    =& -[\mPhi^{\top}\mD_d\mPhi]_i^2|[\vtheta-\tilde{\vtheta}]_i|^2
    +[\vtheta-\tilde{\vtheta}]_i  \left[ \lim_{k\to\infty}\sum_{j=1}^q \lambda_j \gamma \mPhi^{\top}\mD_d\mPhi\mP_{\pi_{\vx^j_k}}\mPhi(\vtheta-\tilde{\vtheta})\right]_i\\
    =& \lim_{k\to\infty}\left( -[\mPhi^{\top}\mD_d\mPhi]_i^2 + \gamma \sum_{j=1}^q\lambda_j [\mPhi^{\top}\mD_d\mP\mPi_{\pi_{\vx^j_k}} \mPhi ]_{i,i} \right) \left|[\vtheta-\tilde{\vtheta}]_i\right|^2\\
    &+ \lim_{k\to\infty} \gamma \sum_{l\in[p]\setminus \{i\}}\sum_{j=1}^q \lambda_j [\mPhi^{\top}\mD_d\mP\mPi_{\pi_{\vx^j_k}}\mPhi]_{i,l}\left[ \vtheta-\tilde{\vtheta} \right]_i\left[ \vtheta-\tilde{\vtheta} \right]_l\\
    =& \lim_{k\to\infty}\sum^q_{j=1}\lambda_j \left(   -[\mPhi^{\top}\mD_d\mPhi]_i^2 + \gamma[\mPhi^{\top}\mD_d\mP\mPi_{\pi_{\vx^j_k}} \mPhi ]_{i,i}   \right)\left\| \vtheta-\tilde{\vtheta}\right\|_{\infty}^2\\
    &  + \lim_{k\to\infty} \sum_{j=1}^q \lambda_j  \gamma \sum_{l\in[p]\setminus \{i\}}[\mPhi^{\top}\mD_d\mP\mPi_{\pi_{\vx^j_k}}\mPhi]_{i,l}\left[ \vtheta-\tilde{\vtheta} \right]_i\left[ \vtheta-\tilde{\vtheta} \right]_l\\
    \leq & \lim_{k\to\infty}\sum^q_{j=1}\lambda_j a_{\min} \left\|\vtheta-\tilde{\vtheta}\right\|^2_{\infty}\\
    =& a_{\min } \left\|\vtheta-\tilde{\vtheta}\right\|^2_{\infty}
\end{align*}
where the second equality follows from simple algebraic decomposition and the last inequality follows from the choice that $|[\vtheta-\tilde{\vtheta}]_i|=\left\| \vtheta-\tilde{\vtheta}\right\|_{\infty}$ and from the definition of $a_{\min}$:
\begin{align*}
    a_{\min}:= \max_{\vx\in\gD_{F_{\mathrm{linear}}}}\max_{i\in [p]}\left(   -[\mPhi^{\top}\mD_d\mPhi]_i^2 + \gamma[\mPhi^{\top}\mD_d\mP\mPi_{\pi_{\vx}} \mPhi ]_{i,i} + \gamma \sum_{l\in[p]\setminus \{i\}} \left| [\mPhi^{\top}\mD_d\mP\mPi_{\pi_{\vx}}\mPhi]_{i,l}\right| \right).
\end{align*}

The third statement (one-sided Lipschitzness of regularized Q-learning) follows from the same logic of second statement.

\end{proof}

\subsection{Proof of Proposition~\ref{prop:q-learning-convergence}}\label{app:sec:prop:q-learning-convergence}
\begin{proof}
    The proof follows from applying the Borkar and Meyn Theorem in Lemma~\ref{lem:borkar-meyn} in Appendix~\ref{app:prelim}. Let us verify the items in Assumption~\ref{borkar_meyn_assumption} in Appendix~\ref{app:prelim}: 
    
    Let us first check item 2 and item 3 of Assumption~\ref{borkar_meyn_assumption} are verifiedWe can see that the ODE counterpart of the Q-learning we consider admits globally asymptotically stable equilibrium point from one-sided liipschitzness in Lemma~\ref{lem:q-learning-osl} and Theorem~\ref{prop:pbe-existence:1}.

    Now, let us verify the remaining items of Assumption~\ref{borkar_meyn_assumption}. Global Lipschitz condition of item 1 follows from the fact that max-operator is a Lipschitz operator. The fourth can be verified using triangle inequalities and fifth item follows from our assumption on the Robbins-Monro step-size~\citep{robbins1951stochastic}.
\end{proof}

\begin{lemma}[Convergence of AVI]\label{lem:convergence-of-avi}
Consider the update in~(\ref{avi}). If~(\ref{suff-cond:avi-convrge:1}) or~(\ref{suff-cond:avi-convrge:2}) holds, then a unique solution of~(\ref{pboe}), say $\vtheta^*$ exists, and $\vtheta_k\to\vtheta^*$
\end{lemma}
\begin{proof}
    Let us consider the condition in~(\ref{suff-cond:avi-convrge:1}). Multiply $\mPhi$ on both sides, and then subtracting $\mPhi\vtheta^*$, we get

    \begin{align*}
     \left\|   \mPhi (\vtheta_{k+1}-\vtheta^*)  \right\|_{\infty} =& \left\| \mPhi(\mPhi^{\top}\mD_d\mPhi)^{-1}\mPhi^{\top}\mD_d(\gamma\mP\mPi_{\pi^g_{\mPhi\vtheta_k}}\mPhi\vtheta_k-\gamma \mP\mPi_{\pi^g_{\mPhi\vtheta^*}}\mPhi\vtheta^*) \right\|_{\infty}\\
     \leq & c \left\|\mPhi\vtheta_k-\mPhi\vtheta^*\right\|_{\infty}
    \end{align*}
    where  $c:= \sup_{\vtheta\in\gD}\gamma \left\| \mPhi (\mPhi^{\top}\mD_{\mu_{\beta_{\mPhi\vtheta}}}\mPhi)^{-1}\mPhi^{\top}\mD\mP\mPi_{\pi_{\mPhi\vtheta}} \right\|_{\infty} <1$, and the second inequality follows from Lemma~\ref{lem:lebourg-mvt-multi-dimension} in the Appendix. Therefore, We have $\left\|  \mPhi (\vtheta_{k+1}-\vtheta^*) \right\|_{\infty}\to 0$. The same argument holds when~(\ref{suff-cond:avi-convrge:2}) holds. 
\end{proof}

\section{MDP examples}\label{sec:mdp_examples}

We define the TD-fixed point for a policy \(\pi\) as \(\vtheta^{\pi} := (\mPhi^{\top} \mD \mPhi - \gamma \mPhi \mD \mP \mPi_{\pi} \mPhi)^{-1} \mPhi^{\top} \mD \mR\). For each greedy policy \(\pi\), if the greedy policy \(\pi^g_{\vtheta^{\pi}}\) induced by the TD-fixed point \(\vtheta^{\pi}\) differs from \(\pi\), then \(\vtheta^{\pi}\) is not a solution to the PBE.



\begin{figure}[ht]
    \centering
\begin{subfigure}[t]{0.23\textwidth}
    \centering
    \includegraphics[height=3cm]{figures/mdp2-avi-not-converge-trajectory.png}
\caption{AVI does not converges. It shows oscillatory behavior}
\end{subfigure}
\hfill
\begin{subfigure}[t]{0.23\textwidth}
    \centering
    \includegraphics[height=3cm]{figures/mdp2-q-converge.png}
\caption{Linear Q-learning converges to a unique solution}
\end{subfigure}
\hfill
\begin{subfigure}[t]{0.23\textwidth}
    \centering
    \includegraphics[height=3cm]{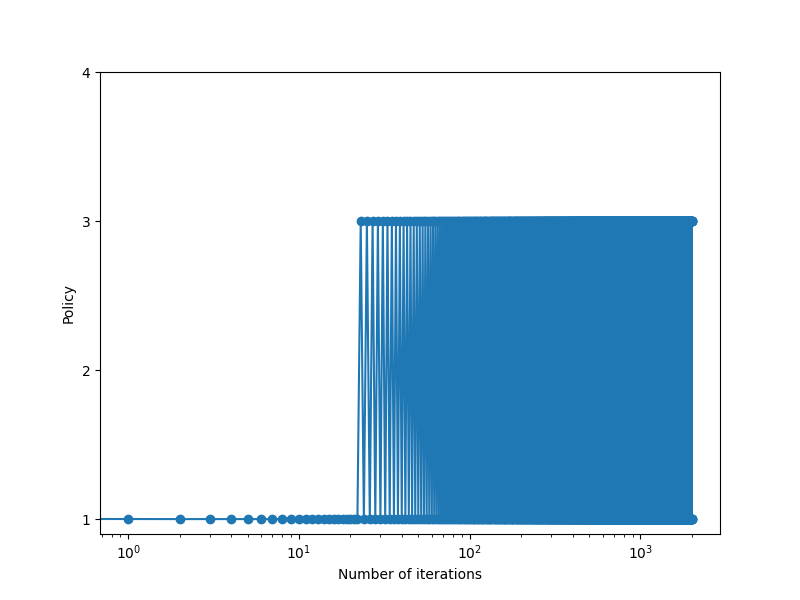}
\caption{Policy change of AVI. For simplicity, each $y$-index, $\{1,2,3,4\}$ corresponds to a deterministic policy. The policy corresponding to $1$ is the optimal policy.}
\end{subfigure}
\hfill
\begin{subfigure}[t]{0.23\textwidth}
    \centering
    \includegraphics[height=3cm]{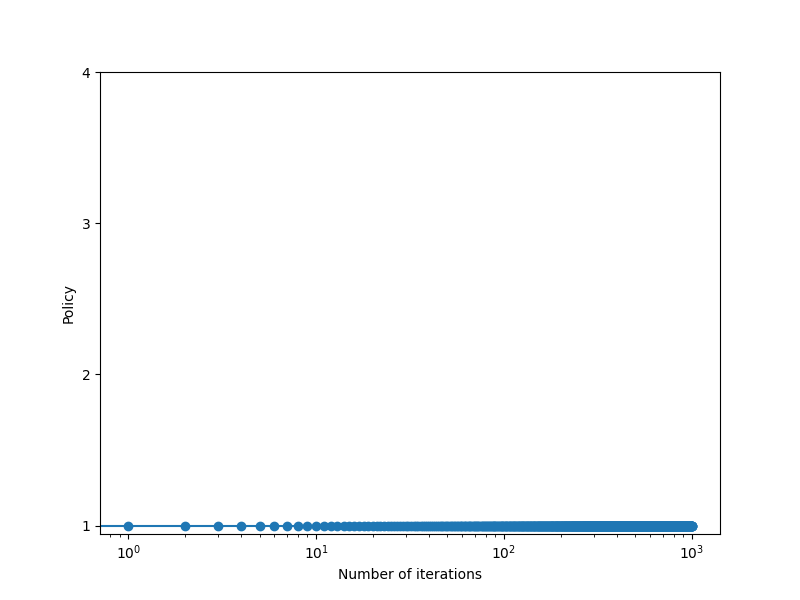}
\caption{Policy change of linear Q-learning.}
\end{subfigure}

\caption{Experimental results on Example~\ref{ex:all-snrdd-but-not-avi-convergence}.}\label{fig:q-learning-converge-not-avi}

\end{figure}

\begin{example}[Q-learning converges but AVI does not]\label{ex:all-snrdd-but-not-avi-convergence}
Consider an MDP with $|\gS| = |\gA| = 2$ and $p = 2$:
\begin{align*}
    \mPhi = \begin{bmatrix}
        0.34 & -0.59\\
        0.25 & -0.16\\
        -0.92 & 0.37\\
        0.83 & 0.19
    \end{bmatrix},\quad \mP = \begin{bmatrix}
        0 & 1 \\
        0.02 &0.98\\
        0.99 & 0.01\\
        0.05 & 0.95
    \end{bmatrix}, \quad \mR = \begin{bmatrix}
        0.3\\
        -0.47\\
        -0.87\\
        -1
    \end{bmatrix},\quad \beta(1\mid 1)=0.96,\quad \beta(1\mid 2)=0.19.
\end{align*}
Then, for any $\pi\in\Omega$, where $\Omega$ is the set of deterministic policies, we can check that $-\mPhi^{\top}\mD_{\mu_{\beta}}\mPhi+\gamma\mPhi^{\top}\mD_{\mu_{\beta}}\mP\mPi_{\pi}\mPhi$ is SNRDD. Therefore, by Theorem~\ref{prop:pbe-existence:1}, there exists a unique solution to PBE, $\vtheta^*\approx \begin{bmatrix}
    -0.67\\
    -1.76
\end{bmatrix} $. Moreover, we can check that $\rho(\gamma(\mPhi^{\top}\mD_{\mu_{\beta}}\mPhi)^{-1}\mPhi^{\top}\mD_{\mu_{\beta}}\mP\mPi_{\pi_{\vtheta^*}}\mPhi)\approx 1.08>1$, and AVI algorithm will not converge. Experimental results are given in Figure~(\ref{fig:avi-q-learning-convergence}) and Figure~(\ref{fig:q-learning-converge-not-avi}).

\end{example}

\begin{figure}[ht]
    \centering
\begin{subfigure}[t]{0.23\textwidth}
    \centering
    \includegraphics[height=3cm]{figures/mdp10-avi-converge-trajectory.png}
\caption{AVI converges to a unique solution}
\end{subfigure}
\hfill
\begin{subfigure}[t]{0.23\textwidth}
    \centering
    \includegraphics[height=3cm]{figures/mdp10-q-not-converge.png}
\caption{Linear Q-learning does not converge}
\end{subfigure}
\hfill
\begin{subfigure}[t]{0.23\textwidth}
    \centering
    \includegraphics[height=3cm]{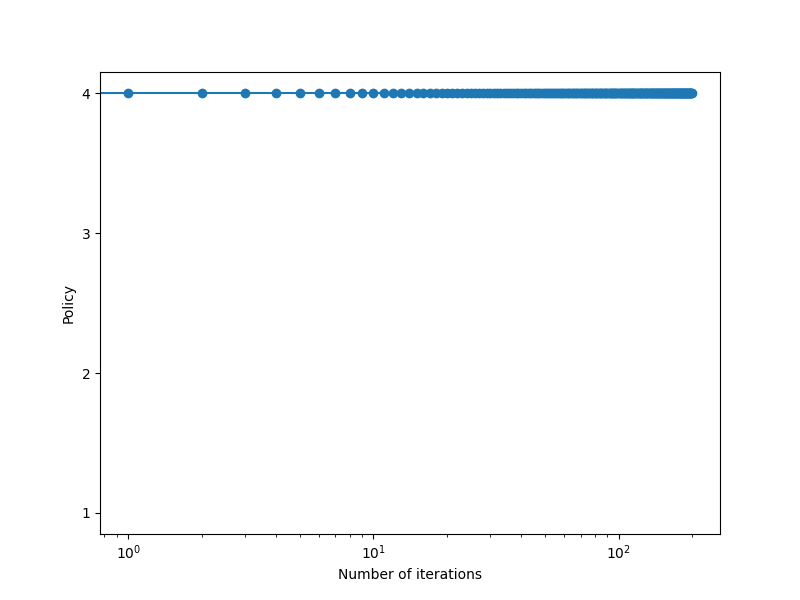}
\caption{Policy change of AVI. For simplicity, each $y$-index, $\{1,2,3,4\}$ corresponds to a deterministic policy. The policy corresponding to $4$ is the optimal policy.}
\end{subfigure}
\hfill
\begin{subfigure}[t]{0.23\textwidth}
    \centering
    \includegraphics[height=3cm]{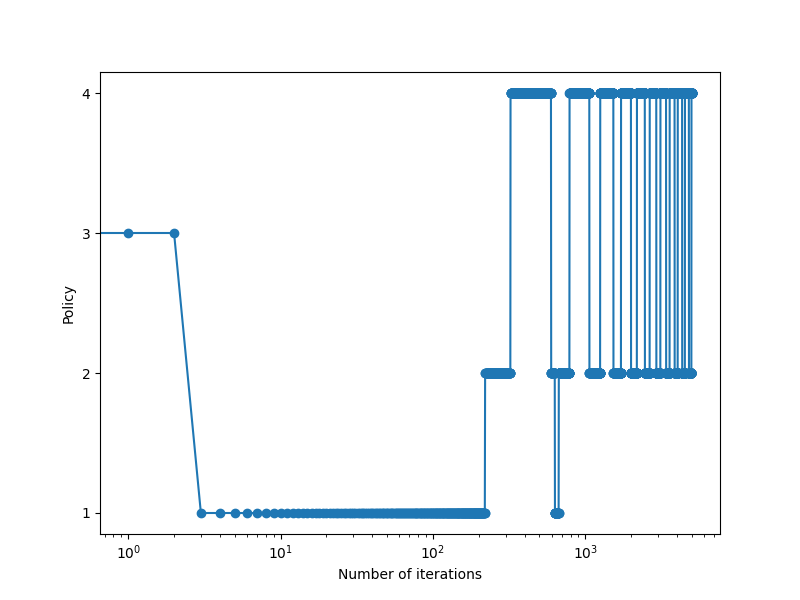}
\caption{Policy change of linear Q-learning.}
\end{subfigure}

\caption{Experimental results on Example~\ref{ex:all-avi-but-hurwitz}.}\label{fig:avi-converge-but-q-learning-not}

\end{figure}

\begin{example}[AVI converges but Q-learning does not ]\label{ex:all-avi-but-hurwitz}

\begin{align*}
    \mPhi =& \begin{bmatrix}
        0.37 & 0.99\\
        0.97 & 1 \\
        -1 & -0.95\\
        -0.77 & 0.19
    \end{bmatrix}
    ,\quad \mP = \begin{bmatrix}
        0.99 & 0.01\\
        0.99 & 0.01\\
        0.89 & 0.11\\
        0.42 & 0.58
    \end{bmatrix},\quad \mR= \begin{bmatrix}
        -0.31\\
        -0.46\\
        -0.35\\
        0.73
    \end{bmatrix},\quad \beta(1\mid 1) = 0.59,\; \beta(1 \mid 2) =0.98.
\end{align*}

One can check that the solution to PBE is $\vtheta^*\approx\begin{bmatrix}
    -1.26\\
    0.89
\end{bmatrix}$. It satisfies~(\ref{suff-cond:avi-convrge:2}), and hence AVI converges. Nonetheless, $-\mPhi^{\top}\mD_{\mu_{\beta}}\mPhi+\gamma\mPhi^{\top}\mD_{\mu_{\beta}}\mP\mPi_{\pi^g_{\vtheta^*}}\mPhi$ is not a Hurwtiz matrix, and therefore, Q-learning does not converge to $\vtheta^*$. The results are shown in Figure~(\ref{fig:avi-q-learning-convergence}) and Figure~(\ref{fig:avi-converge-but-q-learning-not}).

\end{example}

\begin{example}[SNRDD can lead convergence to a point which induces sub-optimal policy]\label{ex:snrdd:local-minima}
    \begin{align*}
        \mPhi = \begin{bmatrix}
            0.13 & 0.09\\
            1 & 0.84\\
            -0.59 & 0.64\\
            -0.94 & -0.28
        \end{bmatrix},\quad \mP = \begin{bmatrix}
            0.99 & 0.01\\
            0.37 & 0.63\\
            0.99 & 0.01\\
            0.99 & 0.01
        \end{bmatrix},\quad \mR= \begin{bmatrix}
            -0.48\\
            0.48\\
            0.41\\
            0.18
        \end{bmatrix},\quad \beta(1\mid 1) = 0.98, \; \beta(1\mid 2)=0.96 
    \end{align*}
    There are two solutions, which are $\vtheta_1^* \approx \begin{bmatrix} -1.26\\-0.27 \end{bmatrix}$ and $\vtheta^*_2\approx  \begin{bmatrix}
        -0.45\\0.98
    \end{bmatrix}$. One can check that  $-\mPhi^{\top}\mD_{\mu_{\beta}}\mPhi+\gamma\mPhi^{\top}\mD_{\mu_{\beta}}\mP\mPi_{\pi^g_{\vtheta^*_1}}\mPhi$ is SNRDD and if we initialize nearby by $\vtheta^*_1$, then the iterate of the Q-learning will converge to $\vtheta^*_1$. Meanwhile, the optimal policy corresponds to the greedy policy induced by $\vtheta^*_2$ whereas $\vtheta^*_1$ induces a sub-optimal policy, i.e, the expected sum of discounted return is lower. The experimental results can be verified in Figure~(\ref{fig:snrdd-sub-optimal-traj}).
\end{example}

\section{$\eps$-greedy solution example}\label{app:sec:eps}

\begin{example}\label{ex:eps-greedy-change-of-solution}
    
Consider a MDP with $|\gS|=1$ and $|\gA|=2$:

\begin{align*}
    \mPhi = \begin{bmatrix}
        0.45\\
        0.79
    \end{bmatrix}, \quad \mP = \begin{bmatrix}
        1\\
        1
    \end{bmatrix},\quad \mPi_{\pi^{\eps}_1} = \begin{bmatrix}
        \eps & 1-\eps
    \end{bmatrix},\quad \mPi_{\pi^{\eps}_2} = \begin{bmatrix}
        1-\eps & \eps
    \end{bmatrix}, \quad \mR= \begin{bmatrix}
        0.5\\
        -0.78
    \end{bmatrix},
\end{align*}

where $\pi^{\eps}_1$ and$\pi^{\eps}_2$ are two different $\eps$-greedy policies. The corresponding stationary distribution of $\mPi_{\pi^{\eps}_1}$ and $\mPi_{\pi^{\eps}_2}$ is $\mD_{\mu_{\pi^{\eps}_1}}=\begin{bmatrix}
    \eps & 0 \\
     0 & 1-\eps
\end{bmatrix}$ and $\mD_{\mu_{\pi^{\eps}_2}}=\begin{bmatrix}
    1-\eps & 0\\
    0      & \eps
\end{bmatrix}$. From Figure~(\ref{fig:eps-solution-change-eps}), we can check that once a critical value is crossed over, then the number of solution changes. 


~\cite{bertsekas2011approximate,young2020understanding} provided examples that the number of solution changes depending on the value of transition probability or reward. Our example differs as the change is determined by the value of $\eps$, which reflects the degree of exploration.
\end{example}

\begin{example}[$\eps$-greedy adds stable unstable solution]\label{ex:eps-unstable}

Consider the following MDP with $|\gS|=1,|\gA|=2$ and $\gamma=0.99$: 

\begin{align*}
    \mPhi =& \begin{bmatrix}
        x\\
        y
    \end{bmatrix},\quad \mP = \begin{bmatrix}
        1\\
        1
    \end{bmatrix},\quad  \mPi_{\pi^{\eps}_1} = \begin{bmatrix}
        1-\eps & \eps
    \end{bmatrix},\quad \mPi_{\pi^{\eps}_2} = \begin{bmatrix}
        \eps & 1 - \eps
    \end{bmatrix},\quad \mR = \begin{bmatrix}
        r_1\\
        r_2
    \end{bmatrix}\\
    \mPi_{\pi^g_1} =& \begin{bmatrix}
        1 & 0
    \end{bmatrix},\quad \mPi_{\pi^g_2} = \begin{bmatrix}
        0 & 1
    \end{bmatrix},\quad \mD_{\mu_{\pi^{\eps}_1}} = \begin{bmatrix}
        1-\eps & 0 \\
        0 & \eps
    \end{bmatrix}, \quad \mD_{\mu_{\pi^{\eps}_2}} = \begin{bmatrix}
        \eps & 0\\
        0 & 1-\eps 
    \end{bmatrix}
\end{align*}

where  $\pi^g_1$ and $\pi^g_2$ represent greedy policies that choose the first and second action, respectively, while $\pi^{\eps}_1$ and $\pi^{\eps}_2$ are the corresponding $\epsilon$-greedy policies, respectively.

Then, we can calculate the following quantities:

\begin{align*}
    \mPhi^{\top}\mD_{\mu_{\pi^{\eps}_1}}\mPhi =& (1-\eps)x^2+\eps y^2, \quad \mPhi^{\top}\mD_{\mu_{\pi^{\eps}_2}}\mPhi = \eps x^2+(1-\eps)y^2, \\
    \mPhi^{\top}\mD_{\mu_{\pi^{\eps}_1}}\mP\mPi_{\pi^g_1} \mPhi=&  \begin{bmatrix}
        x & y 
    \end{bmatrix} \begin{bmatrix}
        1-\eps  & 0\\
        \eps & 0
    \end{bmatrix} \begin{bmatrix}
        x\\
        y
    \end{bmatrix}= (1-\eps)x^2 + +\eps xy ,\\
    \mPhi^{\top}\mD_{\mu_{\pi^{\eps}_2}}\mP\mPi_{\pi^g_2}\mPhi =& \begin{bmatrix}
        x & y 
    \end{bmatrix} \begin{bmatrix}
        0 & \eps\\
        0 & 1-\eps
    \end{bmatrix}\begin{bmatrix}
        x\\
        y
    \end{bmatrix}= (1-\eps)y^2+\eps xy, \\
    \mPhi^{\top} \mD_{\mu_{\pi^{\eps}_1}} \mR =& (1-\eps)x r_1 + \eps y r_2,\quad \mPhi^{\top}\mD_{\mu_{\pi^{\eps}_2}}\mR = \eps x r_1 + (1-\eps)yr_2.
\end{align*}
Now, we can see that 
\begin{align*}
    A_1 =& \mPhi^{\top}\mD_{\mu_{\pi^{\eps}_1}}\mPhi - \gamma  \mPhi^{\top}\mD_{\mu_{\pi^{\eps}_1}} \mP\mPi_{\pi^g_1} \mPhi =   (1-\eps)x^2 +\eps  y^2  - \gamma ((1-\eps)x^2+\eps xy ) \\
    =& \eps( -(1-\gamma) x^2- \gamma xy  + y^2)+ (1 -\gamma) x^2 \\
    A_2 =&  \mPhi^{\top}\mD_{\mu_{\pi^{\eps}_2}}\mPhi   - \gamma   \mPhi^{\top}\mD_{\mu_{\pi^{\eps}_2}}\mP\mPi_{\pi^g_2}\mPhi =  \eps x^2+(1-\eps)y^2-\gamma ((1-\eps)y^2 + \eps xy )\\
    =& \eps(x^2 - \gamma xy - (1-\gamma)y^2)+(1-\gamma)y^2
\end{align*}

Therefore we can now calculate $\theta^{\pi^g_1}$ and $\theta^{\pi^g_2}$, respectively:

\begin{align*}
    \theta^{\pi^g_1} =&  \frac{(1-\eps)xr_1+\eps y r_2}{   \eps( -(1-\gamma) x^2- \gamma xy  + y^2)+ (1 -\gamma) x^2  } ,\\
    \theta^{\pi^g_2} =& \frac{\eps x r_1 + (1-\eps)yr_2}{\eps(x^2 - \gamma xy - (1-\gamma)y^2)+(1-\gamma)y^2}.
\end{align*}

Suppose $y>x>0$ and $r_1,r_2<0$. For $\theta^{\pi_1^g}$ to be a solution, we require $\theta^{\pi^g_1}<0$ which is satisfied if  $A_1>0$. Likewise, $\theta^{\pi^g_2}$ to be a solution, we would require $A_2<0$.

Let $x=0.5$ and $y=1$. Then $A_1= 0.5 \eps + 0.0025$, and for $\eps > -0.005$, $A_1>0$ holds. Therefore, for $\eps\in(0,1)$, $\theta^{\pi^g_1}$ is a solution fo PBE. Meanwhile, $A_2=-0.255\eps +0.01$ and $A_2<0$ holds if $ 0.04< \eps $. Therefore, $\theta^{\pi^g_2}$ becomes a solution of PBE when $ 0.04< \eps $.

Let us discuss the stability of each point in terms of Q-learning. Note that as $A_1>0$, Q-learning will converge to this solution while $A_2<0$, Q-learning will not converge to this solution. 

The optimality of each policy depends on the relative values of $r_1$ and $r_2$. When $r_2 < r_1$, the policy $\pi^g_1$ becomes optimal. Conversely, if $r_1 < r_2$, then the policy $\pi^g_2$ is the optimal policy.

\end{example}

\section{Related Works on linear Q-learning}\label{app:sec:related-works}

This section provides additional literature on linear Q-learning. Several studies have proposed variations of linear Q-learning.~\cite{chen2023target} explored the use of target networks and truncation, while~\cite{maei2010toward,devraj2017zap,carvalho2020new} employed a two-time-scale approach to design a convergent Q-learning algorithm. Although these methods ensure boundedness or convergence, the exact points to which the algorithm converges remain not well understood. In a slightly different setting,~\cite{lu2021convex} explored a linear programming formulation of Q-learning under deterministic transitions. Furthermore,~\cite{chetarget2024} examined Q-learning with a target network in an overparameterized regime, where the number of features exceeds the size of state-action space.

The set of realizable policy by the linear feature set~\citep{lu2018non} is defined as
\begin{align*}
    \left\{ \pi \in \Omega : \pi(s)=arg\max_{a\in\gA} \vphi(s,a)^{\top}\vtheta, \vtheta\in \R^p  \right\}.
\end{align*}
The optimal policy $\pi^*$ may not be in abaove set, and therefore, the solution of PBE might induce only sub-optimal policies.

\section{Pseudo code}\label{sec:algo}

\begin{algorithm}[h]
\caption{(regularized) Q-learning with linear function approximation}
\begin{algorithmic}[1]

\State Initialize $\vtheta_0\in \mathbb{R}^p, \;\eta\in\R$.

\For{iteration step $k\in \{0,1,\ldots\}$}
\State Observe $s_k,a_k\sim d(\cdot)$, $s_k^{\prime}\sim \gP(\cdot\mid s_k,a_k)$,  and $r_k=r(s_k,a_k,s_{k}^{\prime})$.
\State Update parameters according to
\begin{align*}
\vtheta _{k + 1} = \vtheta _k + \alpha _k \vphi(s_k,a_k)(r_k+\gamma \max_{a\in\gA}\vphi^{\top}(s_{k}^{\prime},a)\vtheta_k-\vphi(s_k,a_k)^{\top}\vtheta_k - \eta\vtheta_k).
\end{align*}
\EndFor
\end{algorithmic}
\label{algo:1}
\end{algorithm}

\begin{algorithm}[h]
\caption{Deterministic (regularized) Q-learning with linear function approximation}
\begin{algorithmic}[1]
\State Initialize $\vtheta_0\in \mathbb{R}^p, \;\eta\in\R, d\in\Delta^{\gS\times\gA}$.
\For{iteration step $k\in \{0,1,\ldots\}$}
\State 
\begin{align*}
    \vtheta_{k+1}= \vtheta _k + \alpha _k (\mPhi^{\top}\mD_d\mR+\gamma\mPhi^{\top}\mD_d\mP\mPi_{\pi^g_{\vtheta_k}}\mPhi\vtheta_k- \eta\vtheta_k).
\end{align*}
\EndFor
\end{algorithmic}
\label{algo:deterministic-q}
\end{algorithm}









\end{document}